\documentclass{article}
\usepackage[utf8]{inputenc}

\title{Quantifying VIO Uncertainty}
\author{Stephanie Tsuei, Stefano Soatto}
\date{January 2023}

\usepackage{graphicx}
\usepackage{amsmath}
\usepackage{amssymb}
\usepackage{booktabs}
\usepackage{xcolor}
\usepackage{multirow}
\usepackage{bbm}
\usepackage{subfigure}
\usepackage{soul}

\usepackage[pagebackref,breaklinks,colorlinks]{hyperref}

\usepackage{fullpage}

\begin{document}

\maketitle

\begin{abstract}
We compute the uncertainty of XIVO, a monocular visual-inertial odometry system based on the Extended Kalman Filter, in the presence of Gaussian noise, drift, and attribution errors in the feature tracks in addition to Gaussian noise and drift in the IMU. Uncertainty is computed using Monte-Carlo simulations of a sufficiently exciting trajectory in the midst of a point cloud that bypass the typical image processing and feature tracking steps. We find that attribution errors have the largest detrimental effect on performance. Even with just small amounts of Gaussian noise and/or drift, however, the probability that XIVO's performance resembles the mean performance when noise and/or drift is artificially high is greater than 1 in 100.
\end{abstract}

\section{Introduction}
Our companion paper \cite{tsuei_feature_2023} observed that the feature tracks commonly used in visual inertial odometry algorithms are affected by drift, noise, and attribution errors. This manuscript follows up on the companion and investigates the effect of drift, noise, and attribution errors on state estimation performance and uncertainty of XIVO, our in-house visual-inertial odometry system previously known as Corvis \cite{jones_visual-inertial_2011, hernandez_observability_2015}, in simulation.\footnote{\url{https://github.com/ucla-vision/xivo}} Using a simulation rather than a benchmark dataset of real-world data allows us to quantify the individual effects of each rather than a mixture of all three. It also enables Monte-Carlo trials to calculate uncertainty without using the ergodicity assumption in \cite{tsuei_2021}. Since our previous work in \cite{tsuei_2021} has shown that covariance matrices estimated by XIVO are not accurate, all mentions of ``uncertainty" or ``covariance" refers to sample covariances calculated using Monte-Carlo trials.

Most existing works on visual-inertial odometry benchmark performance on a real-world dataset of motion sequences consisting of IMU data and RGB images. When using current real-world datasets, there is little opportunity to benchmark uncertainty, as each motion sequence is only collected once.\footnote{If the ergodicity assumption holds, the methods in \cite{tsuei_2021} can be used to compute uncertainty using real data.} Simulated cause-and-effect on our in-house monocular visual-inertial system allows us to study the effects of Gaussian noise, drift, and attribution errors individually on a general implementation. In real-world data, all these effects are tangled together.

Section \ref{sec:xivo_features_calib_prelims} details the exact definitions and equations used in this chapter. Section \ref{sec:xivo_features_calib_experiment} gives details about the experiment. This chapter then ends with some concluding remarks.

\section{Preliminaries}
\label{sec:xivo_features_calib_prelims}

Our notation is consistent with \cite{murray_mathematical_1994} and past works describing the algorithm implemented in XIVO \cite{jones_visual-inertial_2011}, \cite{hernandez_observability_2015}. A detailed description of XIVO and its equations are given in \cite{tsuei_xivodocpdf_2023}.

Let $T$ denote the length of a trajectory in timesteps. Timestamps are denoted with the variable $t$. The state $\mathbf x(t) \in \mathbb R^9$ contains a rotation vector, translation vector, and velocity. Let $\mathbf{\hat x}(t)$ denote the estimated state and $\mathbf e(t) = \mathbf{\hat x}(t) - \mathbf x(t)$ denote the error state.

Our Monte-Carlo trials contain $N$ runs each, indexed by $n$. $\mathbf x_n(t), \mathbf e_n(t)$ are the state and error state at time $t$ in run $n$. The \textbf{sample covariance} at each timestep is given by
\begin{equation}
    \Sigma(t) = \frac{1}{N-1}\sum_{n=1}^N \mathbf e_n(t) \mathbf e_n(t)^T.
\label{eq:samplecov}
\end{equation}
The \textbf{mean sample covariance} over an entire trajectory is
\begin{equation}
    \bar \Sigma = \frac{1}{T} \sum_{t=0}^T \Sigma(t)
    \label{eq:meansamplecov}
\end{equation}

Although not a part of standard metrics for evaluating SLAM systems \cite{sturm_benchmark_2012}, errors in \emph{scale} are the most sensitive to small perturbations in the input data. In other words, the error in translation and linear velocity are state-dependent and well-aligned with the ground-truth translation vector. This magnitude of these errors will vary widely depending on the exact imperfections in the input data. The \textbf{scale factor} $\rho$ of an estimated trajectory is the mean value of the ratio of the norms of the estimated translation $\hat T_{sb}(t)$ and ground-truth translation $T_{sb}(t)$, after both trajectories are centered around their centroids:
\begin{equation}
    \rho = \frac{1}{T} \sum_{t=0}^T \frac{\| \hat{\tilde T}_{sb}(t)\|}{\| \tilde T_{sb}(t)\|}
\label{eq:scale_def}
\end{equation}
    where
\begin{equation} 
\begin{aligned}
    \tilde T_{sb}(t) &= T_{sb}(t) - \frac{1}{T} \sum_{\tau=0}^T T_{sb}(\tau) \\
    \hat{\tilde T}_{sb}(t) &= \hat T_{sb}(t) - \frac{1}{T} \sum_{\tau=0}^T \hat T_{sb}(\tau).
\end{aligned}
\end{equation}
When $\rho < 1$, the estimated trajectory is ``smaller'' than the ground-truth trajectory. When $\rho > 1$, the estimated trajectory is ``larger'' than the ground-truth trajectory. To avoid division-by-zero errors when $T_{sb}$ is close to the origin, we only include timesteps when $ \| T_{sb}(t) \| > 0.1$ when  computing $\rho$. For a set of $N$ Monte-Carlo trials indexed by $n$, we can compute a set of scale factors $\rho_n$ using equation \eqref{eq:scale_def} and plot their distribution.

\subsection{On Observability and Identifiability for Monocular VIO}

Observability is a property of a model that is assumed to contain a dynamic state, inputs, and outputs. Informally, a model is \emph{observable} if given the inputs and outputs, the trajectory of the dynamic state can be exactly determined. A model is \emph{unknown-input observable} if the state of the model can be exactly determined even if a subset of the inputs are not known. If the model contains calibration parameters to be estimated online, then the model is \emph{identifiable} if the value of the calibration parameters and the trajectory of the dynamic state can be determined from the inputs and outputs. Parameter identifiability requires a \emph{sufficiently exciting} input, i.e. if the input is not varied enough then it is possible that multiple values of the calibration parameters can satisfy the model dyanmics and output equations. What makes an input sufficiently exciting depends on the model.

For linear time-invariant systems, the definition of a sufficiently exciting input is well-known. The condition is called \emph{persistent excitation}. Various equivalent definitions can be found in \cite{astrom_numerical_1965}, \cite{soderstrom_identifiability_1976}, and \cite{willems_note_2005}.

The exact conditions for sufficient excitation of nonlinear systems is not well understood and is still an active area of study. An early result states that nonlinear systems are locally identifiable if their linearization about an equilibrium point is identifiable \cite{grewal_identifiability_1976} --- then the requirement for parameter identifiability is the same as for a linear time-invariant system, persistent excitation. Examples of recent papers on the topic are \cite{padoan_geometric_2017, verrelli_nonanticipating_2020, tomei_enhanced_2022}. In texts on systems identification and adaptive control, the requirements for sufficient excitation can be derived from LaSalle's invariance theorem and Barbalat's Lemma \cite{lavretsky_robust_2012}. Texts on system identification and adaptive control typically focus on \emph{linear-in-parameters} nonlinear systems, or other assumed relationships between the dynamics and the parameters.

VIO models are not covered by the literature on systems identification and adaptive control.  There are many previous works about the observability and identifiability of VIO models; each paper uses a slightly different model. The definition of sufficient excitation therefore varies from work to work. For example in \cite{jones_visual-inertial_2011}, IMU measurements are outputs and the inputs to the model, linear jerk and angular acceleration, are modeled as noise independent of the state. IMU biases are Brownian motion. In \cite{yang_degenerate_2019}, the IMU measurements are modeled as inputs, rather than outputs, and the state contains an extra time-calibration parameter. IMU input noise is independent of the state. In \cite{hernandez_observability_2015}, IMU measurements are inputs to the model,  IMU input noise is not independent of the state, there is no extra time-calibration parameter, and IMU biases are slowly-varying unknown inputs rather than noise. The model implemented in XIVO and used in these experiments is the one examined in \cite{hernandez_observability_2015}.

\cite{hernandez_observability_2015} proves that the VIO model under examination is identifiable if and only if IMU bias rates are zero or exactly known. Otherwise, there exist multiple trajectories that can satisfy the output equations. The multiple trajectories are, however, a bounded set; the size of the bounded set is proportional to the IMU bias rates and inversely proportional to the \emph{minimum excitation} of the angular velocity, angular acceleration, angular jerk, and linear jerk. The minimum excitation of a 3D signal $f(t)$ over a time interval $t \in \mathcal I$ is defined as:
\begin{equation}
m(f: \mathcal{I}) = \inf_{\|x \|=1} \left ( \sup_{t \in \mathcal I} \| f(t) \times x \| \right ).
\label{eq:minimum_excitation}
\end{equation}
In other words, the minimum excitation of $f(t)$ is determined by the (arbitrary) direction in 3D space with the smallest maximum value. If there is a direction that is not covered at all (e.g. no angular jerk in the $z$-direction), then the minimum excitation is zero, and the bounded set of possible trajectories in \cite{hernandez_observability_2015} is actually unbounded. Sufficient excitation therefore means that angular velocity, angular acceleration, angular jerk, and linear jerk each cover 3-DOF. A randomly generated 3D trajectory will almost surely meet the requirement.

Of course, the model and the state estimation algorithm are different. A state estimation algorithm may fail to find the correct state of an observable model with a sufficiently exciting input. Conversely, it is also possible for a state estimation algorithm to produce the original state even if the model is not observable or the input is not sufficiently exciting.

\section{Experiment}
\label{sec:xivo_features_calib_experiment}

We simulate IMU and visual inputs to our in-house VIO system, XIVO. Inputs are perturbed with Gaussian noise, drift, and random attribution errors so that we may measure the effect of each on Absolute Trajectory Error (ATE), Relative Pose Error (RPE), sample covariance $\Sigma(t)$ \eqref{eq:samplecov}, and scale factor $\rho$ \eqref{eq:scale_def}.

\subsection{The Trajectory Studied}
\begin{figure}
    \centering
    \includegraphics[width=3in]{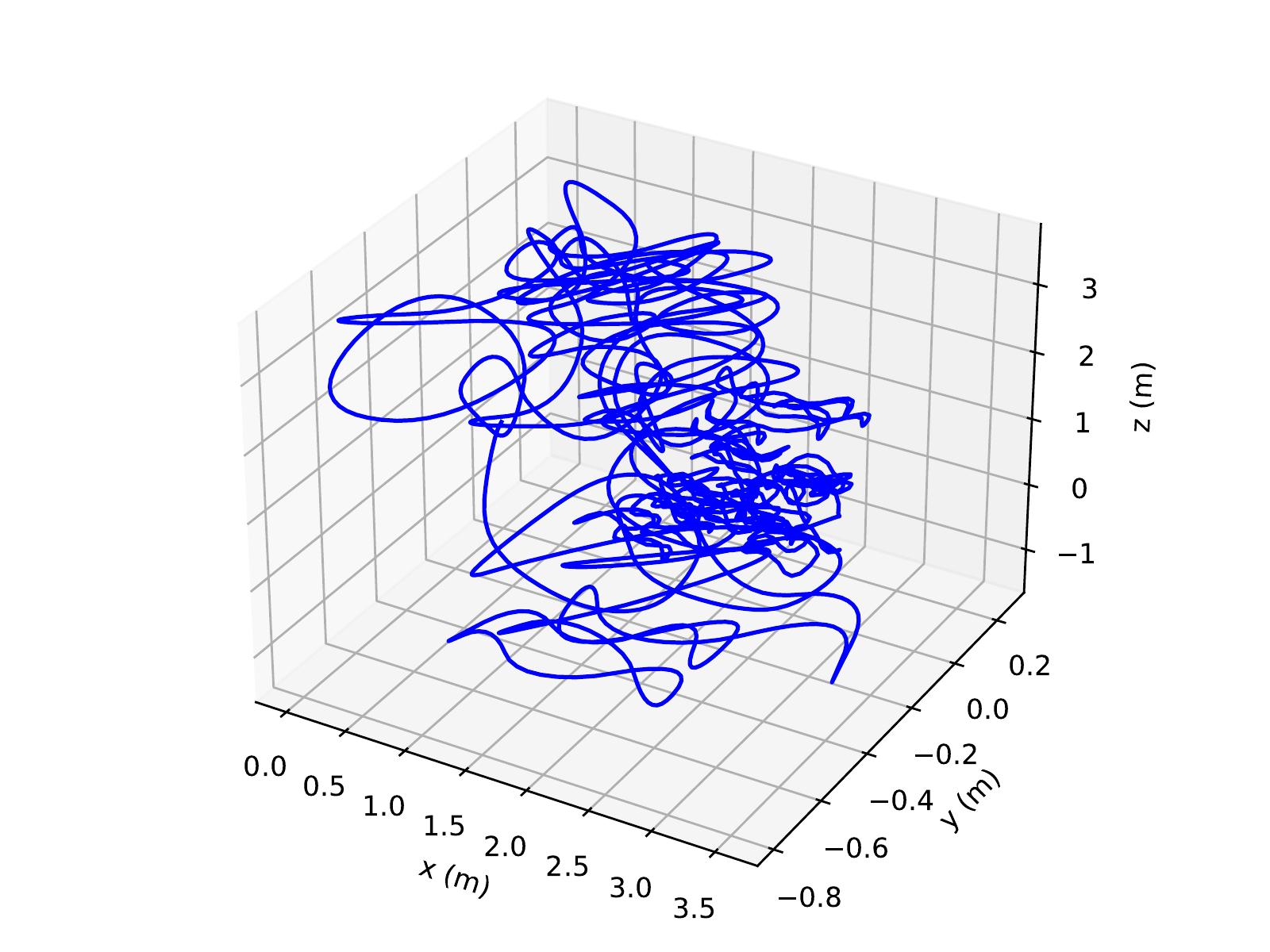}
    \includegraphics[width=3in]{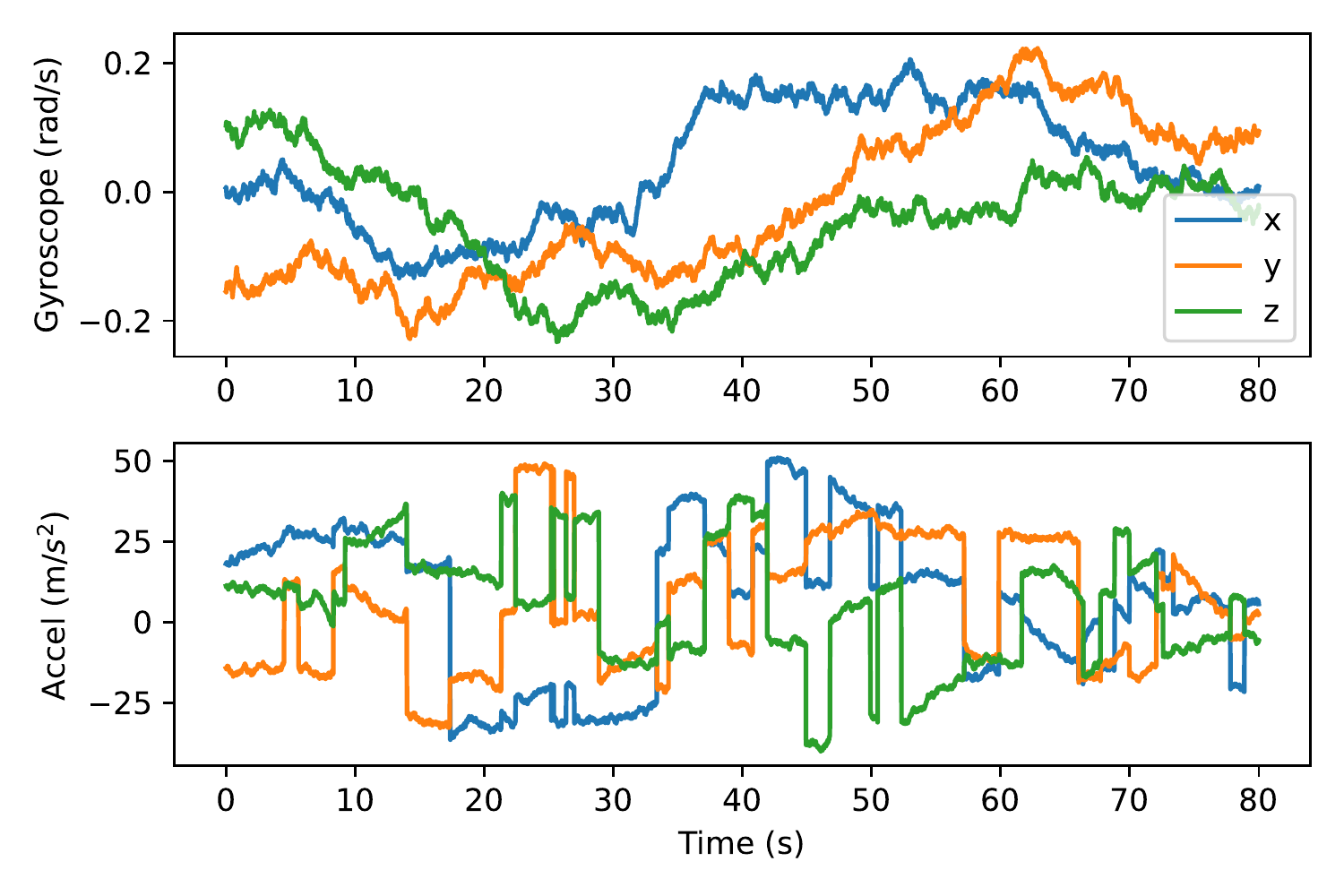}
    \caption{\textbf{The Brownian motion trajectory.} Linear acceleration and angular velocity are modeled as Brownian motion. Translation is plotted in the left figure in 3D. The linear acceleration and angular velocity inputs, in the body frame, are plotted in the right figure. Sudden jumps in the acceleration input correspond to instances when the trajectory hits a boundary condition in the spatial frame.}
    \label{fig:brownian_motion_curve}
\end{figure}

Our experiment focuses on a randomly generated trajectory that is guaranteed to be sufficiently exciting and pictured in Figure \ref{fig:brownian_motion_curve}. In the randomly generated trajectory, linear acceleration $\alpha_{sb}^s$ and angular velocity $\omega_{sb}^s$ in the spatial frame are modeled as Brownian motion with acceleration drift $\sigma_{\alpha} = 0.1$ m and angular velocity drift $\sigma_{\omega} = 0.001$ rad/s. A random amount of drift is added to $\alpha_{sb}^s$ and $\omega_{sb}^s$ with every IMU measurement, at 400Hz. $\alpha_{sb}^s$ and $\omega_{sb}^s$ are then transformed into the body-frame inputs, $\alpha_{sb}^b$ and $\omega_{sb}^b$. Boundary conditions on linear velocity $v_{sb}$ are set to $[ -3, -1, -1]$ m/s and $[3, 1, 1]$ m/s. Boundary conditions on translation $T_{sb}$ are set to $[ -6, -3, -3]$ m and $[ 6, 3, 3]$ m. If the simulated ground-truth state in the spatial frame hits either boundary, the sign of that component of linear acceleration is flipped. Boundary conditions on all components of rotation vector $w_{sb}$ are set to $-\pi$ and $\pi$. The total length of the trajectory is 127.76m over 80 seconds.

Although randomly generated Brownian motion curves, such as the trajectory in Figure \ref{fig:brownian_motion_curve}, are guaranteed to be sufficiently exciting, the frequent and non-smooth changes in rotation and acceleration make convergence of an EKF and feature depth initialization (see \cite{tsuei_xivodocpdf_2023}) difficult due to lack of parallax for triangulation. %

\subsection{Configuration}

\begin{figure}
    \centering
    \includegraphics[width=\textwidth]{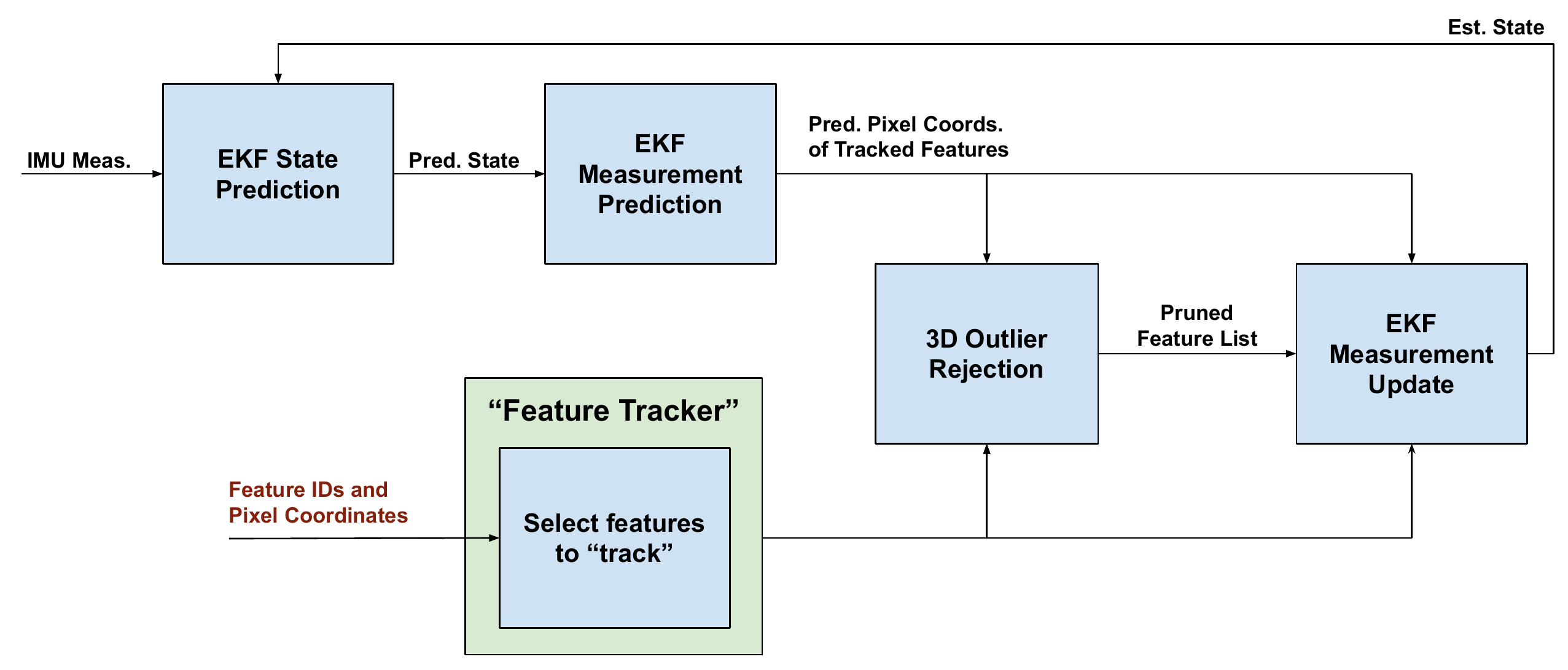}
    \caption{\textbf{XIVO's Configuration for Monte-Carlo Experiments.} For Monte-Carlo experiments, XIVO's typical feature tracker was replaced with simple bookkeeping software. Loop closure, an optional component, was not used. The typical configuration of XIVO is given in Figure 1 of \cite{tsuei_xivodocpdf_2023}.}
    \label{fig:xivo_sim_config}
\end{figure}

The configuration of XIVO in Monte-Carlo experiments is given in Figure \ref{fig:xivo_sim_config}. More details about the configuration are given in the paragraphs below.

\paragraph{IMU Simulation.} We simulate an IMU with similar specifications to the Intel RealSense d435i, a low-cost device. The accelerometer and gyroscope produce measurements at 400 Hz. IMU noise levels are held constant with accelerometer noise $\sigma_a = 1e^{-4}$ m/s$^2/\sqrt{\text{Hz}}$  and gyroscope noise $\sigma_g = 1e^{-5}$ radians/s/$\sqrt{\text{Hz}}$. IMU bias is initially zero, but drifts with parameters $\sigma_{b_a} = 3e^{-4}$ m/s$^2 / \sqrt{\text{Hz}}$ and $\sigma_{b_g} = 5e^{-6}$ radians/s/$\sqrt{\text{Hz}}$.

\paragraph{Vision Simulation.} So that we may have complete control over feature position errors and attribution, we skip the typical image processing step and instead directly feed pixel measurements of an attributed point cloud into XIVO at a rate of 25Hz. The point cloud consists of 1000 randomly generated points uniformly located in a box; all Monte-Carlo trials use the same set of generated features. At each vision timestep Visibility is calculated using a camera with no distortion and the following intrinsics:
\begin{equation}
    K = \begin{bmatrix}
        275 & 0 & 320 \\
        0 & 275 & 240 \\
        0 & 0 & 1
    \end{bmatrix}
    \label{eq:intrinsics}
\end{equation}
Observed features fall in and out of view during motion. 

\paragraph{Extrinsics.} Due to the development of specialized visual-inertial calibration, such as Kalibr, we assume that camera intrinsics, camera-IMU timestamp alignment, camera-IMU extrinsics, and accelerometer-gyroscope alignment are known. As in \cite{jones_visual-inertial_2011}, we ``remove'' camera-IMU extrinsics from the state by setting the initial covariance of that portion of the state to a value $\sim 1e^{-10}$. All other calibration quantities are removed from the state through compile-time switches.

\paragraph{XIVO Configuration.} A bare-bones ``feature tracker'' is configured to ``track'' and initialize depth estimates of 250 - 500 visible features. If there are more than 500 features, they will be ignored. Feature depths are initialized with a combination of triangulation minimizing angular reprojection errors \cite{Lee_2019_ICCV} and subfiltering. There is no loop closure in these experiments. The EKF in XIVO uses 60 tracked features in its state. The features selected for state estimation are those with the most confident estimate of depth (see \cite{tsuei_xivodocpdf_2023} for more details).

\subsection{Experiment Parameters}

\paragraph{Gaussian Noise.} Let $\sigma_p$ be the standard deviation of noise added to feature tracks and let $\bar \sigma_p$ be the standard deviation of noise used by the EKF when creating state estimates. We test values of $\sigma_p \in \{ 0.25,\allowbreak 0.50,\allowbreak 0.75,\allowbreak 1.00,\allowbreak 1.25,\allowbreak 1.50,\allowbreak 1.75,\allowbreak 2.00,\allowbreak 2.25,\allowbreak 2.50 \}$ pixels and set $\bar \sigma_p = \sigma_p$.

\paragraph{Drift.} 
To simulate drift, we associate with each feature $j$ a bias $b^j_p(t) \in \mathbb R^2$, with unit of pixels. When feature $j$ is first detected, $b^j_p(t)$ is initialized to $[0,0]^T$. With each frame, $b^j_p(t)$ evolves as $b^j_p(t+1) = b^j_p(t) + B$ where $B \sim \mathcal N(0, \sigma_b)$. At time $t$, the XIVO receives a simulated tracker measurement of $\pi(X_c(t)) + b^j_p(t)$, where $X^j_c(t) \in \mathbb R^3$ is the location of feature $j$ in the camera frame at time $t$. We test $\sigma_b \in \{ 0.001, 0.005, 0.01, 0.05, 0.1, 0.5 \}$ and two different values for the assumed Gaussian noise uncertainty in the EKF, $\bar \sigma_p = 0.50$ and $\bar \sigma_p = 1.00$. Over 25 frames (one simulated second), this corresponds to an average drift of zero pixels for all values of $\sigma_b$, with standard deviations of $\{ 0.005, 0.025, 0.05, 0.25, 0.5, 2.5\}$ pixels.

\paragraph{Attribution Errors.} To create attribution errors, we swap the measurements of $\eta$ percent of uniformly randomly selected visible features each frame before passing feature tracks to XIVO. The features assigned attribution errors are always those that are ``tracked'' by the bare-bones feature tracker, but may or may not be used in state estimation by the EKF. XIVO is configured to use Mahalanobis Gating to reject outlier measurements, such as those caused by misattributions. Since measurement errors caused by randomly swapping visible features are catastrophic, Mahalanobis Gating prevents the outlier measurements from becoming part of the EKF's measurement update. Therefore, the real effect of our random misattributions is that the lifetime of feature tracks are prematurely cut short. We test with $\eta=0, 1, 2.5, 5, 7.5, 10, 20, 30, 40$ percent.

\subsection{Results}

\begin{figure}
    \centering
    \subfigure[Absolute Trajectory Error]{\includegraphics[width=6in]{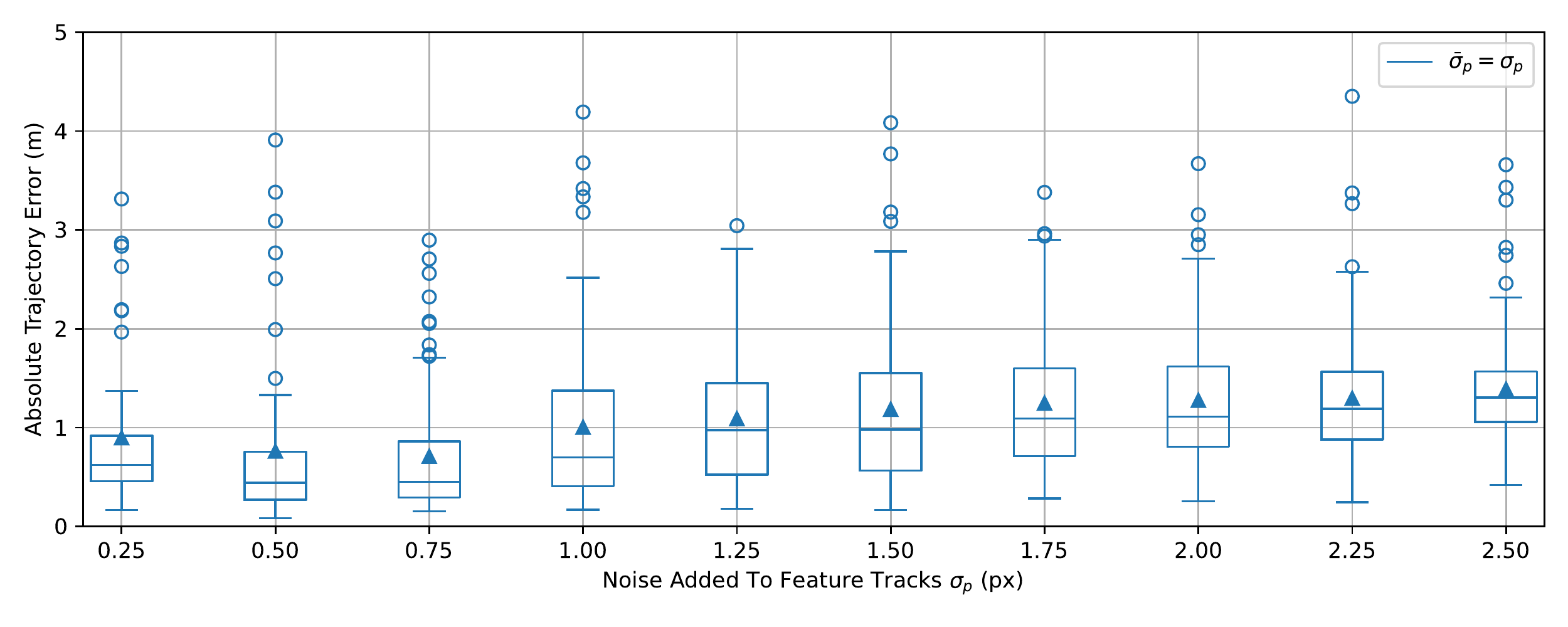}}
    \subfigure[Relative Pose Error]{\includegraphics[width=6in]{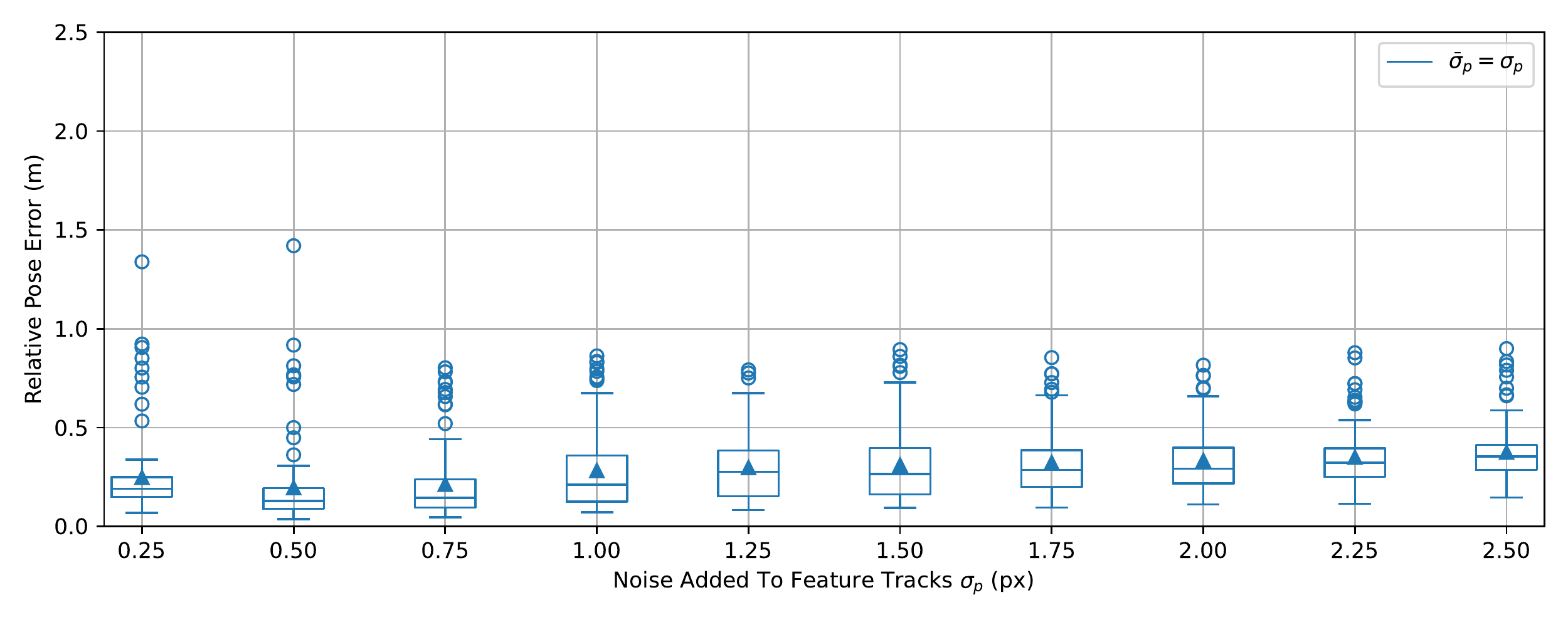}}
    \caption{\textbf{$\bar \sigma_p = \sigma_p$: Performance decreases with Gaussian Noise.} Each box-and-whisker illustrates the distribution of Absolute Trajectory Error (top) and Relative Pose Error (bottom) over 100 Monte-Carlo trials. Boxes extend from the first to the third quartile. Medians are lines in the boxes, means are triangles. Whiskers extend the box by 1.5x the inter-quartile range. All other points are plotted as ``fliers''. Mean and median error increase with noise for all $\bar \sigma_p = \sigma_p \geq 0.50$. The performance is lower for $\sigma_p = 0.25$ than for $\sigma_p = 0.50$ because $\bar \sigma_p = 0.25$ is too small to capture uncertainties due to poor feature initialization in Brownian motion in addition to Gaussian noise.}
    \label{fig:gaussian_noise_performance}
\end{figure}

\begin{figure}
    \centering
    \includegraphics[width=6in]{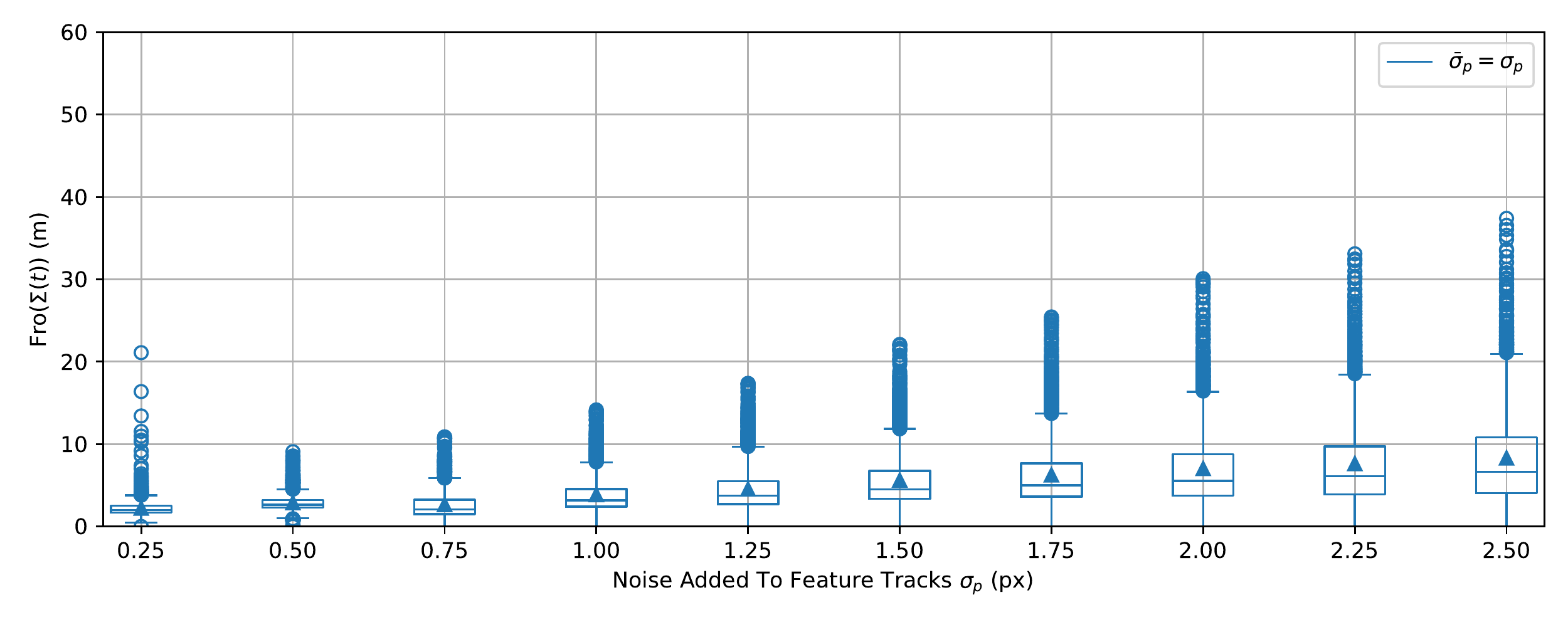}
    \caption{\textbf{$\bar \sigma_p = \sigma_p$: Gaussian Noise leads to larger sample covariances.} Each box-and-whisker illustrates the distribution of sample covariance (eq. \eqref{eq:samplecov}) computed using 100 Monte-Carlo trials. Boxes extend from the first to the third quartile. Medians are lines in the boxes, means are triangles. Whiskers extend the box by 1.5x the inter-quartile range. All other points are plotted as ``fliers''. As the amount of noise increases, so does the sample covariance and the variation in sample covariance.}
    \label{fig:gaussian_noise_covariance_norm}
\end{figure}

\begin{figure}
    \centering
    \includegraphics[width=6in]{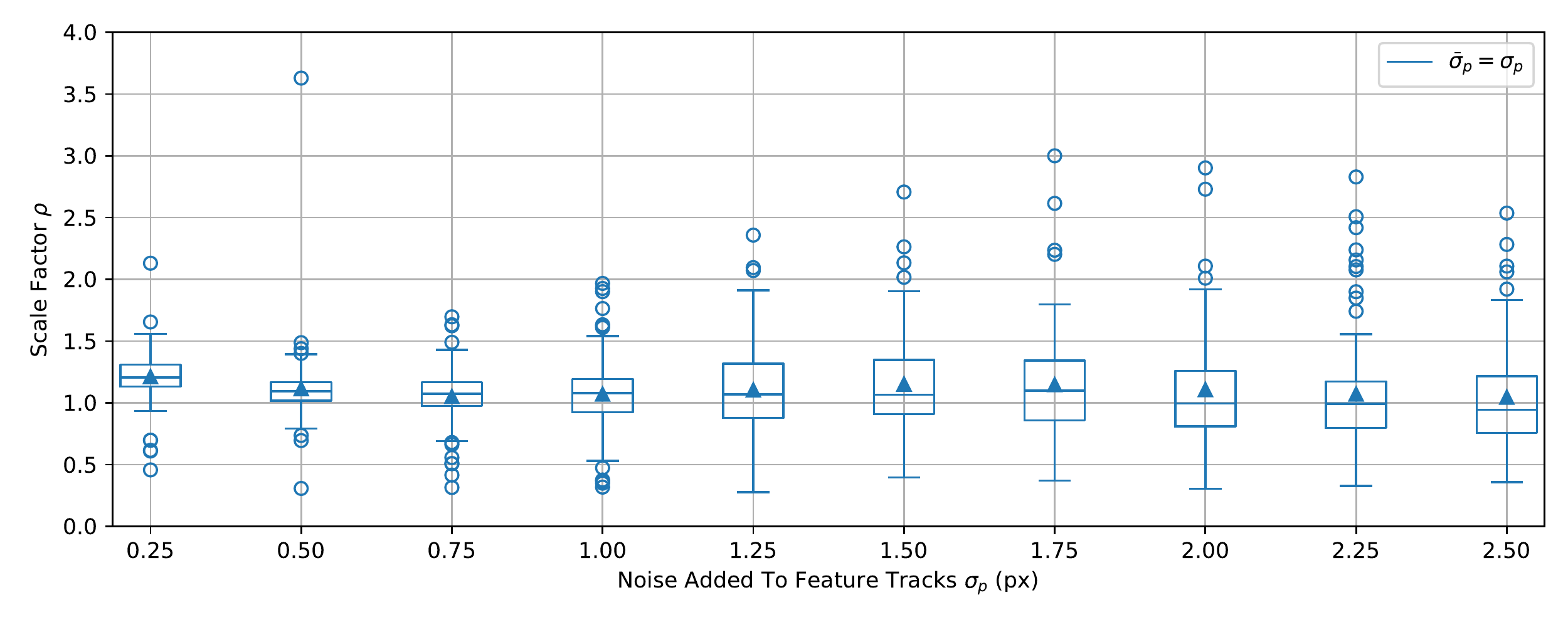}
    \caption{\textbf{$\bar \sigma_p = \sigma_p$: Mean and variation of scale factor $\rho$ is a nonlinear function of $\sigma_p$.} Each box-and-whisker illustrates the distribution of $\rho$ computed using 100 Monte-Carlo trials. Boxes extend from the first to the third quartile. Medians are lines in the boxes, means are triangles. Whiskers extend the box by 1.5x the inter-quartile range. All other points are plotted as ``fliers''. Generally, we see that although there is no trend in the mean or median scale, the variation in scale generally increases with $\sigma_p$. Scale estimates are relatively poor for $\sigma_p = 0.25$ because $\bar \sigma_p = 0.25$ is too small to capture uncertainties due to poor feature initialization in addition to Gaussian noise.}
    \label{fig:gaussian_noise_scale_distributions}
\end{figure}

\begin{figure}
    \centering
    \subfigure[Absolute Trajectory Error]{\includegraphics[width=6in]{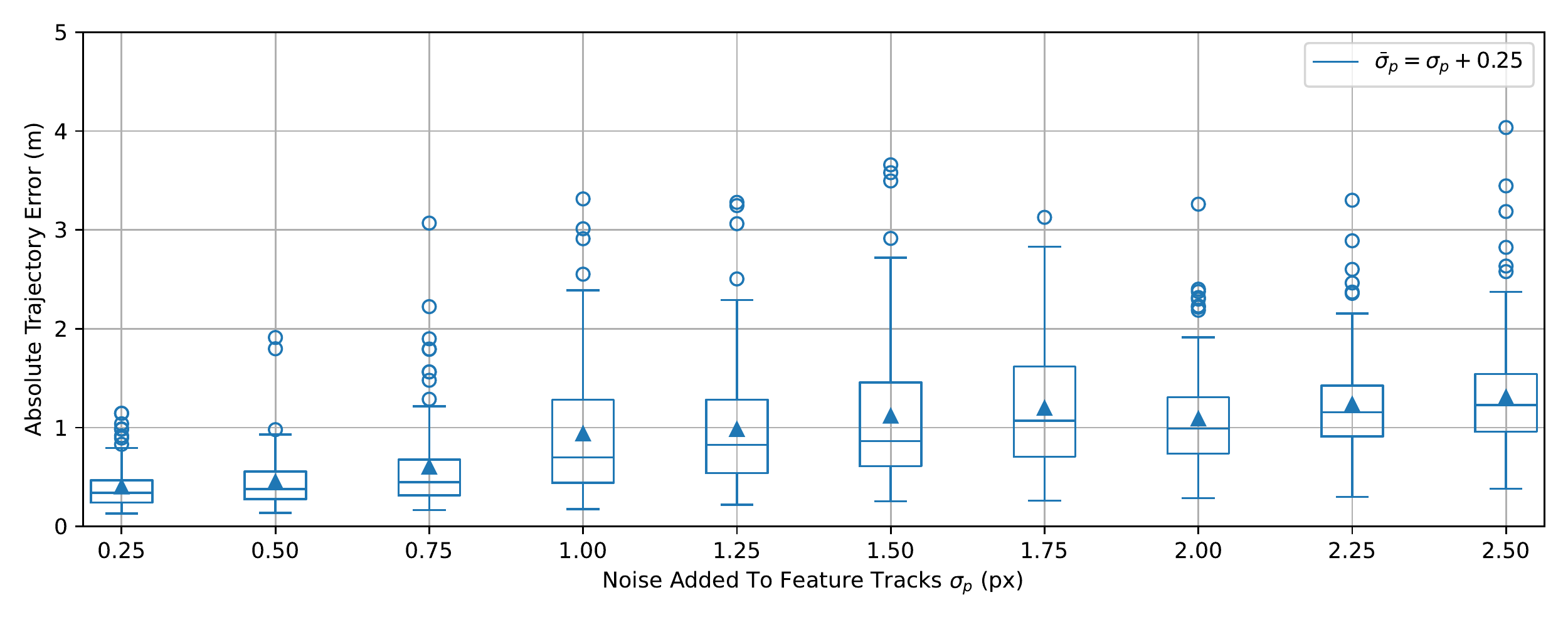}}
    \subfigure[Relative Pose Error]{\includegraphics[width=6in]{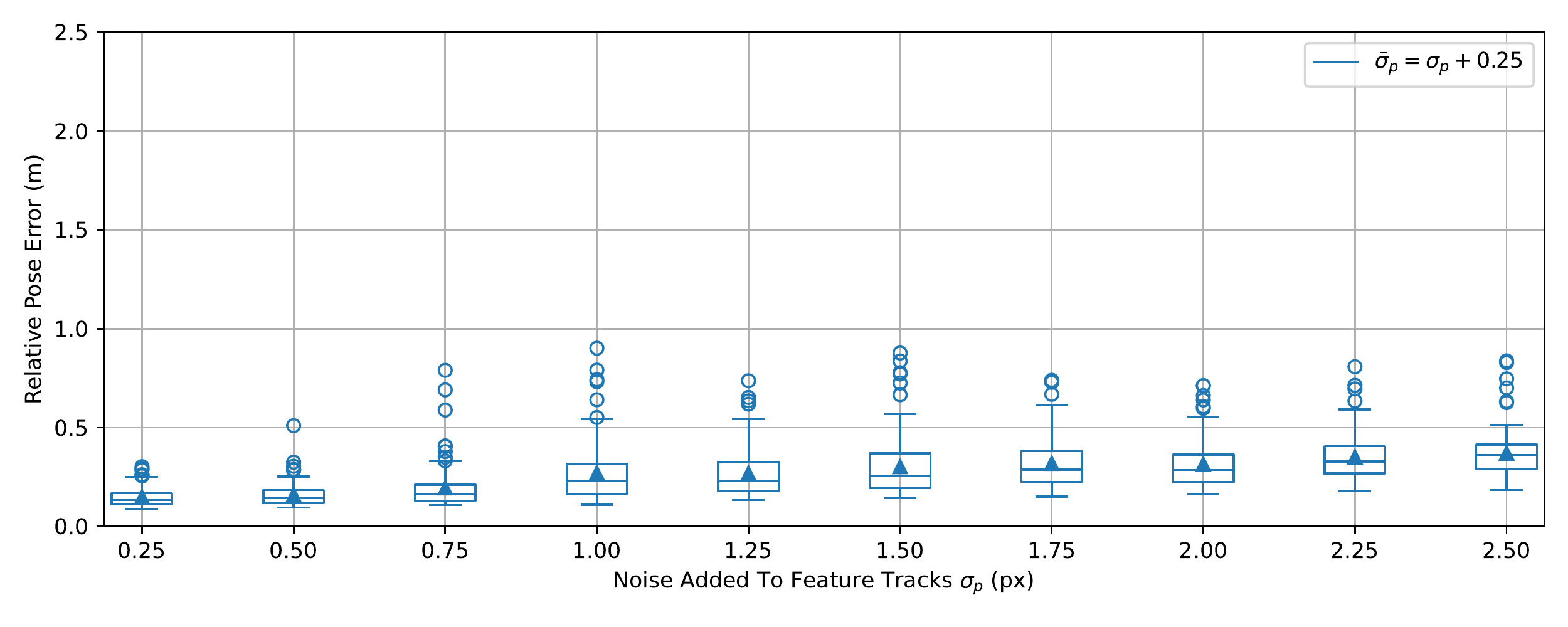}}
    \caption{\textbf{$\bar \sigma_p = \sigma_p + 0.25$: Performance decreases with Gaussian Noise.} Each box-and-whisker illustrates the distribution of Absolute Trajectory Error (top) and Relative Pose Error (bottom) over 100 Monte-Carlo trials. Boxes extend from the first to the third quartile. Medians are lines in the boxes, means are triangles. Whiskers extend the box by 1.5x the inter-quartile range. All other points are plotted as ``fliers''. Mean and median error increase with noise for all $\bar \sigma_p = \sigma_p \geq 0.50$.}
    \label{fig:gaussian_noise_performance2}
\end{figure}

\begin{figure}
    \centering
    \includegraphics[width=6in]{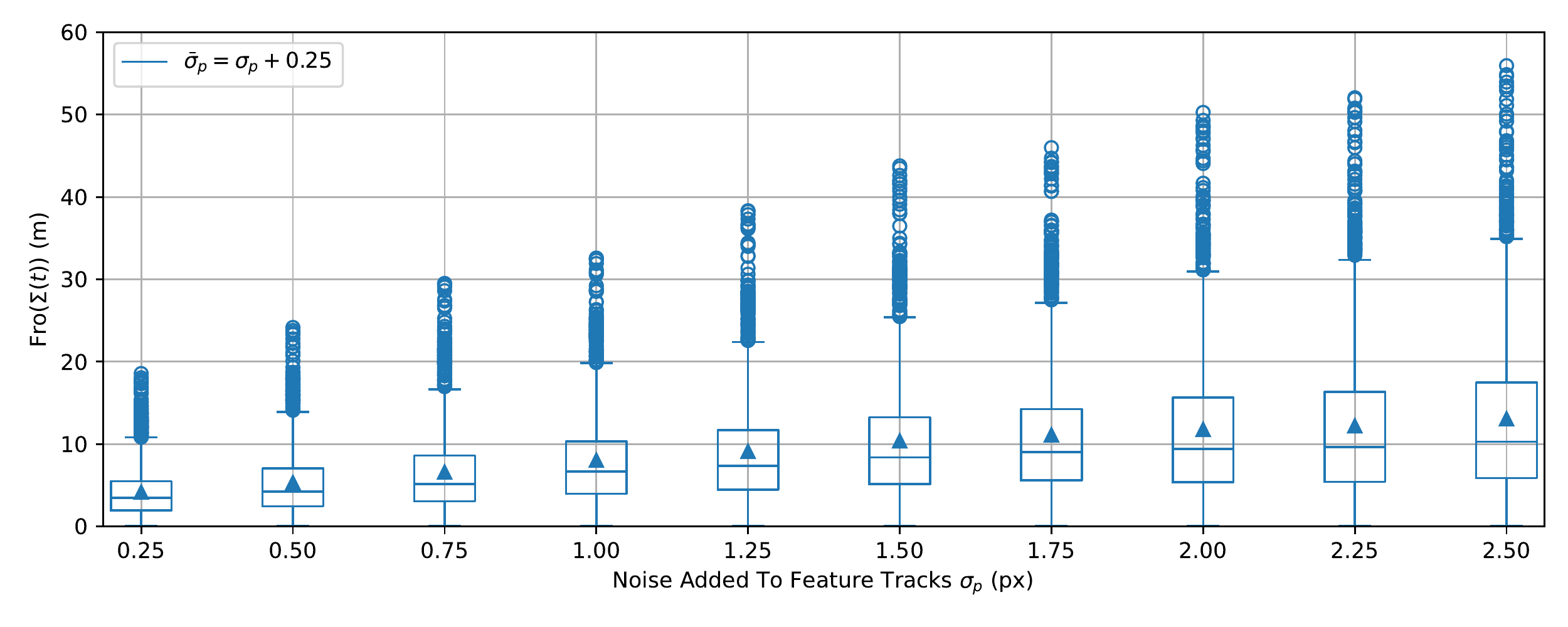}
    \caption{\textbf{$\bar \sigma_p = \sigma_p + 0.25$: Gaussian Noise leads to larger sample covariances.} Each box-and-whisker illustrates the distribution of mean sample covariance (eq. \eqref{eq:samplecov}) computed using 100 Monte-Carlo trials. Boxes extend from the first to the third quartile. Medians are lines in the boxes, means are triangles. Whiskers extend the box by 1.5x the inter-quartile range. All other points are plotted as ``fliers''. As the amount of noise increases, so does the sample covariance and the variation in sample covariance.}
    \label{fig:gaussian_noise_covariance_norm2}
\end{figure}

\begin{figure}
    \centering
    \includegraphics[width=6in]{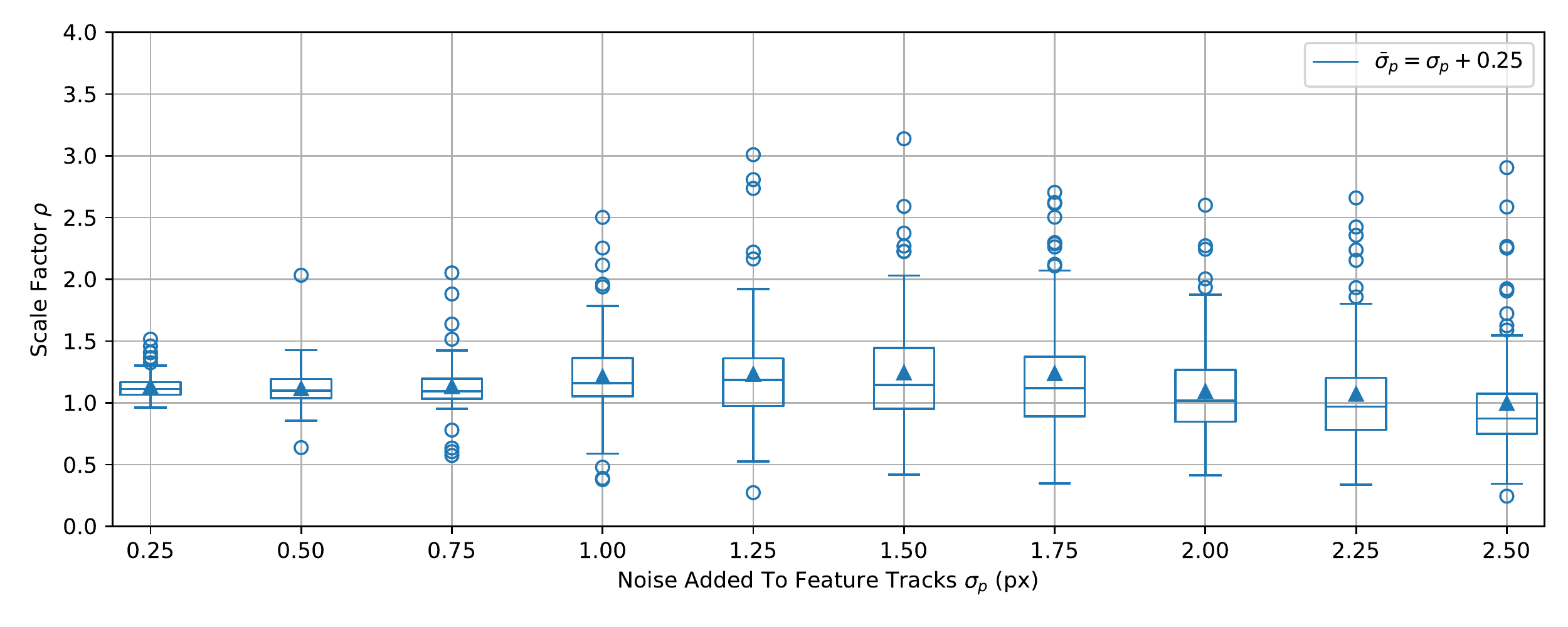}
    \caption{\textbf{$\bar \sigma_p = \sigma_p + 0.25$: Mean and variation of scale factor $\rho$ is a nonlinear function of $\sigma_p$.} Each box-and-whisker illustrates the distribution of $\rho$ computed using 100 Monte-Carlo trials. Boxes extend from the first to the third quartile. Medians are lines in the boxes, means are triangles. Whiskers extend the box by 1.5x the inter-quartile range. All other points are plotted as ``fliers''. Generally, we see that although there is no trend in the mean or median scale, the variation in scale generally increases with $\sigma_p$.}
    \label{fig:gaussian_noise_scale_distributions2}
\end{figure}

\begin{figure}
    \centering
    \subfigure[Absolute Trajectory Error]{\includegraphics[width=6in]{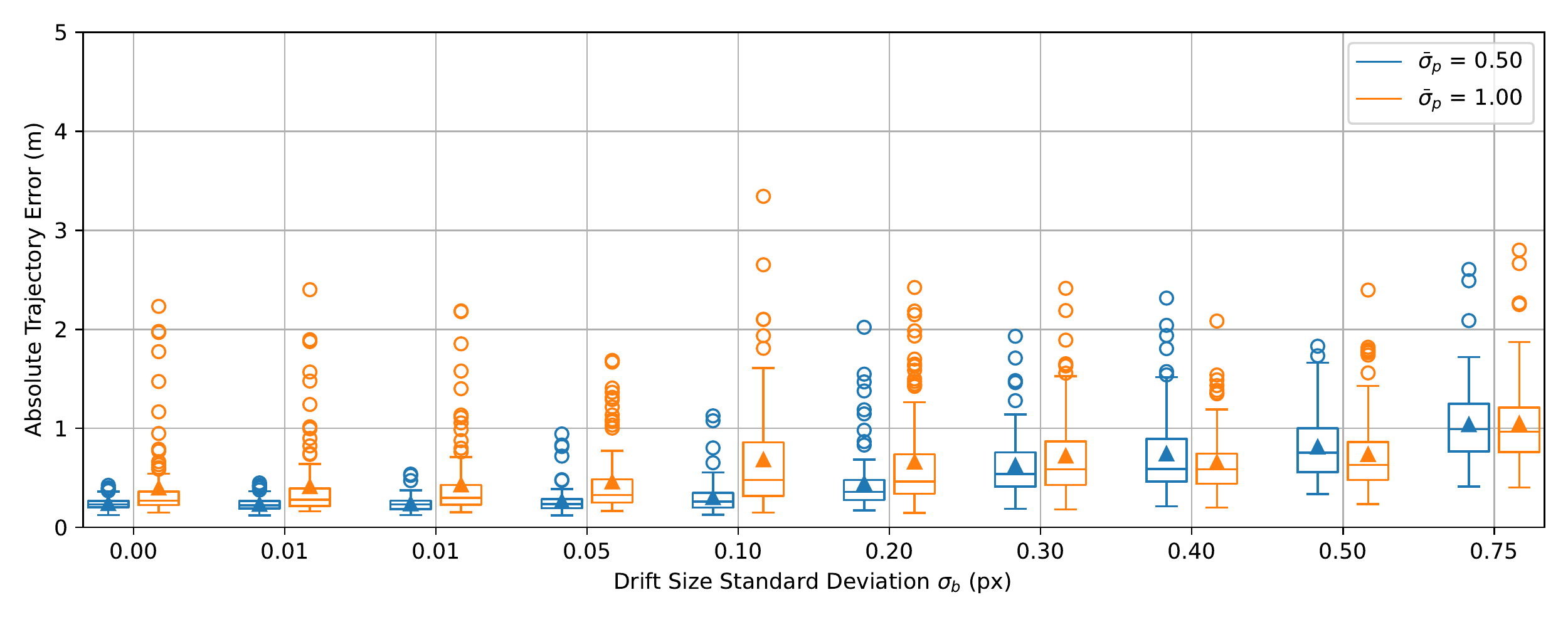}}
    \subfigure[Relative Pose Error]{\includegraphics[width=6in]{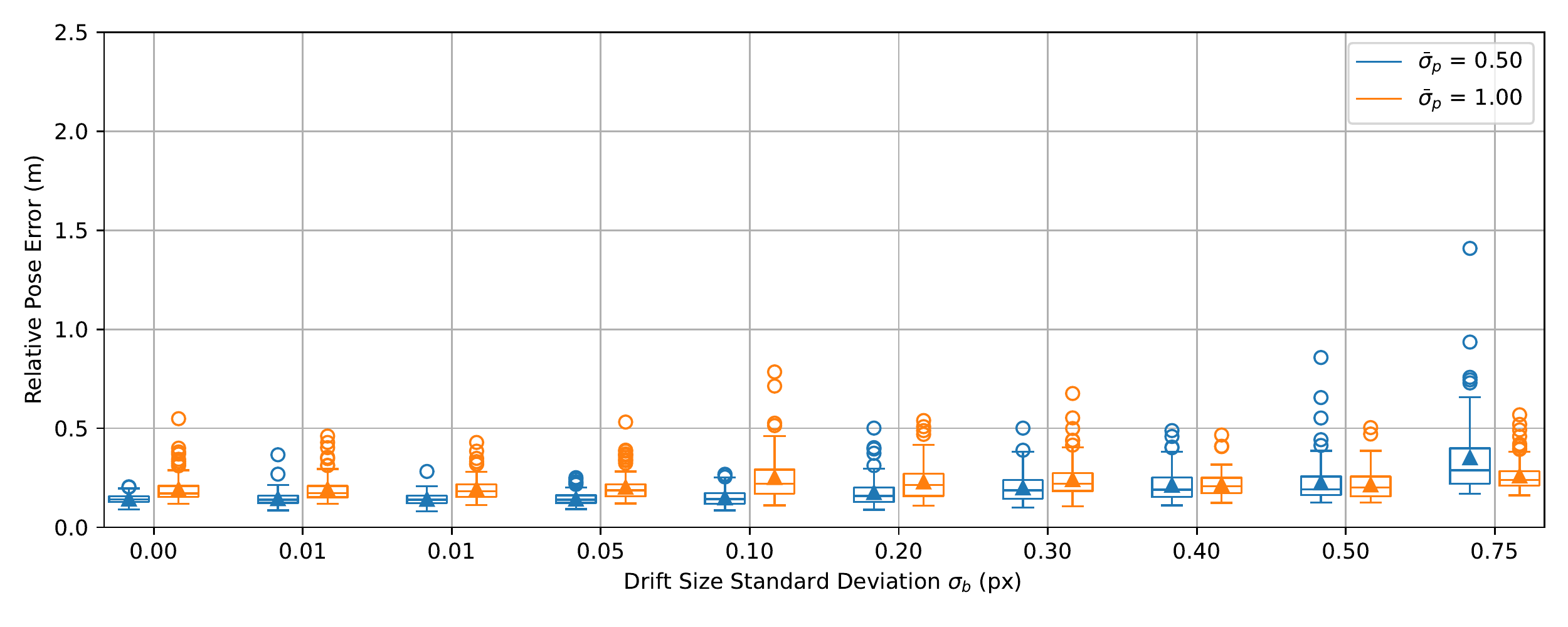}}
    \caption{\textbf{Drift increases estimation error and variation in estimation error.}  Each box-and-whisker illustrates the distribution of ATE and RPE computed using 100 Monte-Carlo trials. Boxes extend from the first to the third quartile. Medians are lines in the boxes, means are triangles. Whiskers extend the box by 1.5x the inter-quartile range. All other points are plotted as ``fliers''. The mean and median performance error creeps upwards with the drift $\sigma_b$ for both $\bar \sigma_p = 0.50$ and $\bar \sigma_p = 1.00$. Performance is better when $\bar \sigma_p = 0.50$ for smaller amounts of drift; for values of $\sigma_b \ge 0.40$, $\bar \sigma_p = 1.00$ produces better performance.}
    \label{fig:drift_performance}
\end{figure}

\begin{figure}
    \centering
    \includegraphics[width=6in]{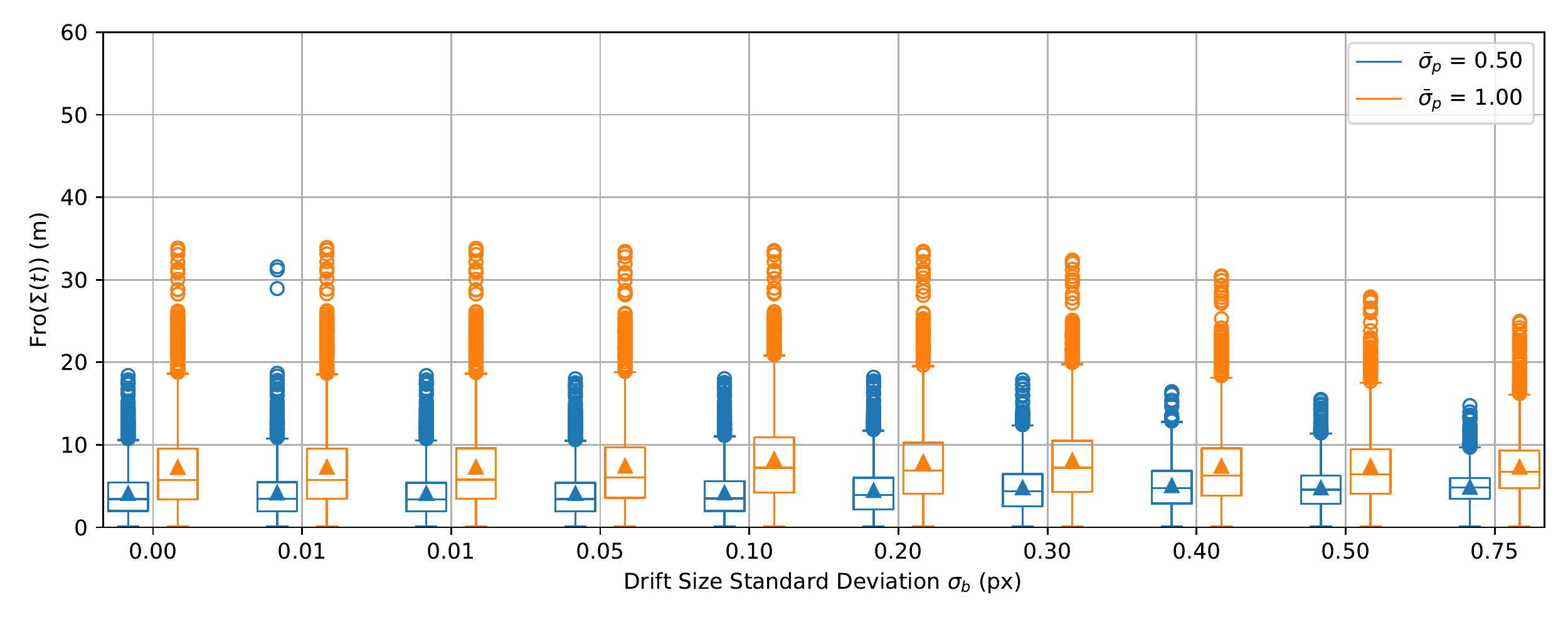}
    \caption{\textbf{Drift only slightly increases state uncertainty.} Each box-and-whisker illustrates the distribution of sample covariance (eq. \eqref{eq:samplecov}) computed using 100 Monte-Carlo trials. Boxes extend from the first to the third quartile. Medians are lines in the boxes, means are triangles. Whiskers extend the box by 1.5x the inter-quartile range. All other points are plotted as ``fliers''. Mean and median covariance size increases with drift. For both $\bar \sigma_p = 0.50$ and $\bar \sigma_p = 1.00$, the mean and median values of drift creep slightly upwards with increasing values of $\bar \sigma_b$.}
    \label{fig:drift_covariance_norm}
\end{figure}

\begin{figure}
    \centering
    \includegraphics[width=6in]{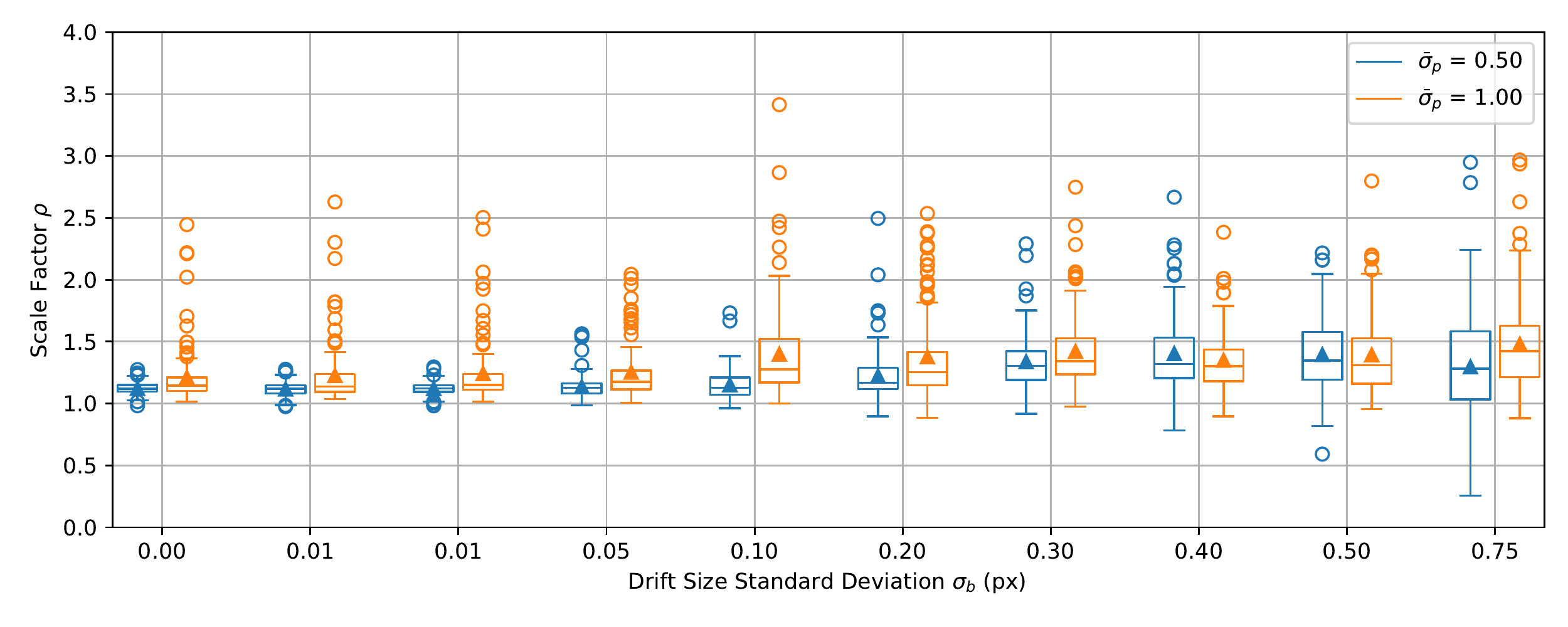}
    \caption{\textbf{Drift increases both bias and uncertainty in scale.}  Each box-and-whisker illustrates the distribution of scale factor (eq. \eqref{eq:scale_def}) computed using 100 Monte-Carlo trials. Boxes extend from the first to the third quartile. Medians are lines in the boxes, means are triangles. Whiskers extend the box by 1.5x the inter-quartile range. All other points are plotted as ``fliers''.}
    \label{fig:drift_scale_distributions}
\end{figure}

\begin{figure}
    \centering
    \subfigure[Absolute Trajectory Error]{\includegraphics[width=6in]{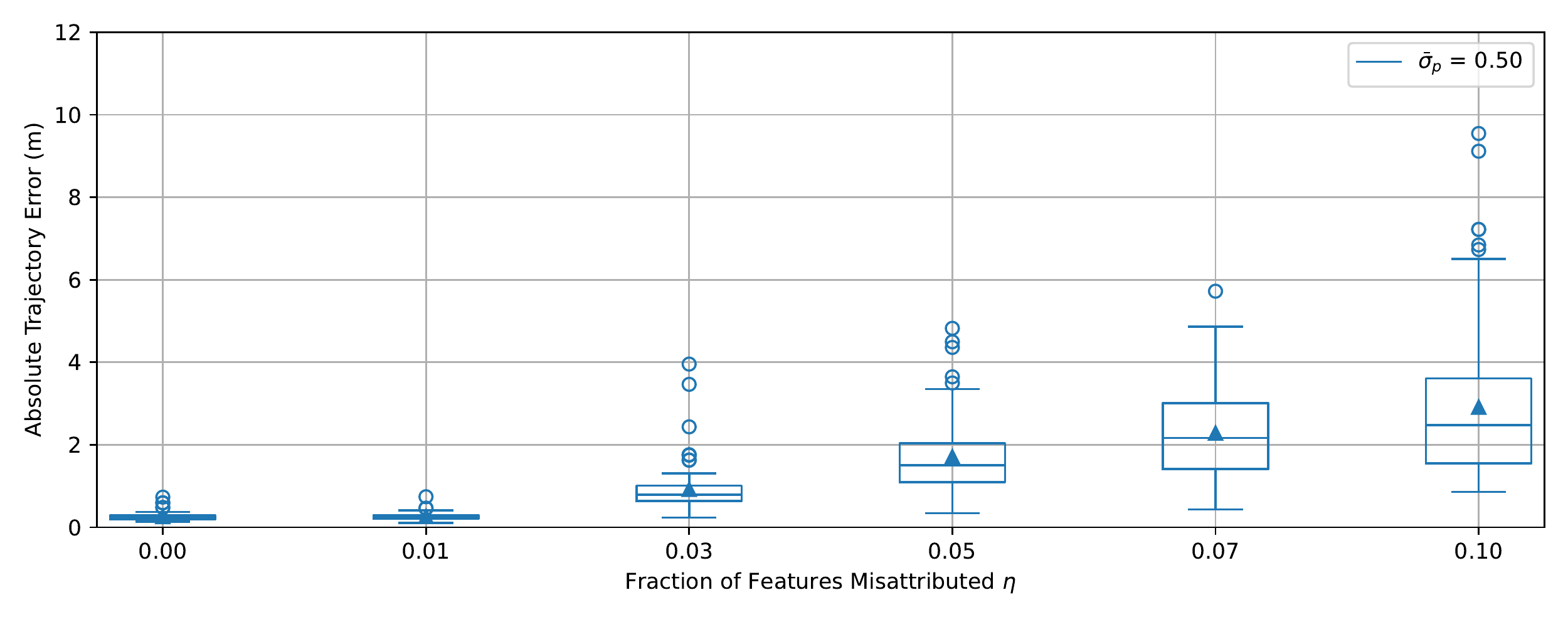}}
    \subfigure[Relative Pose Error]{\includegraphics[width=6in]{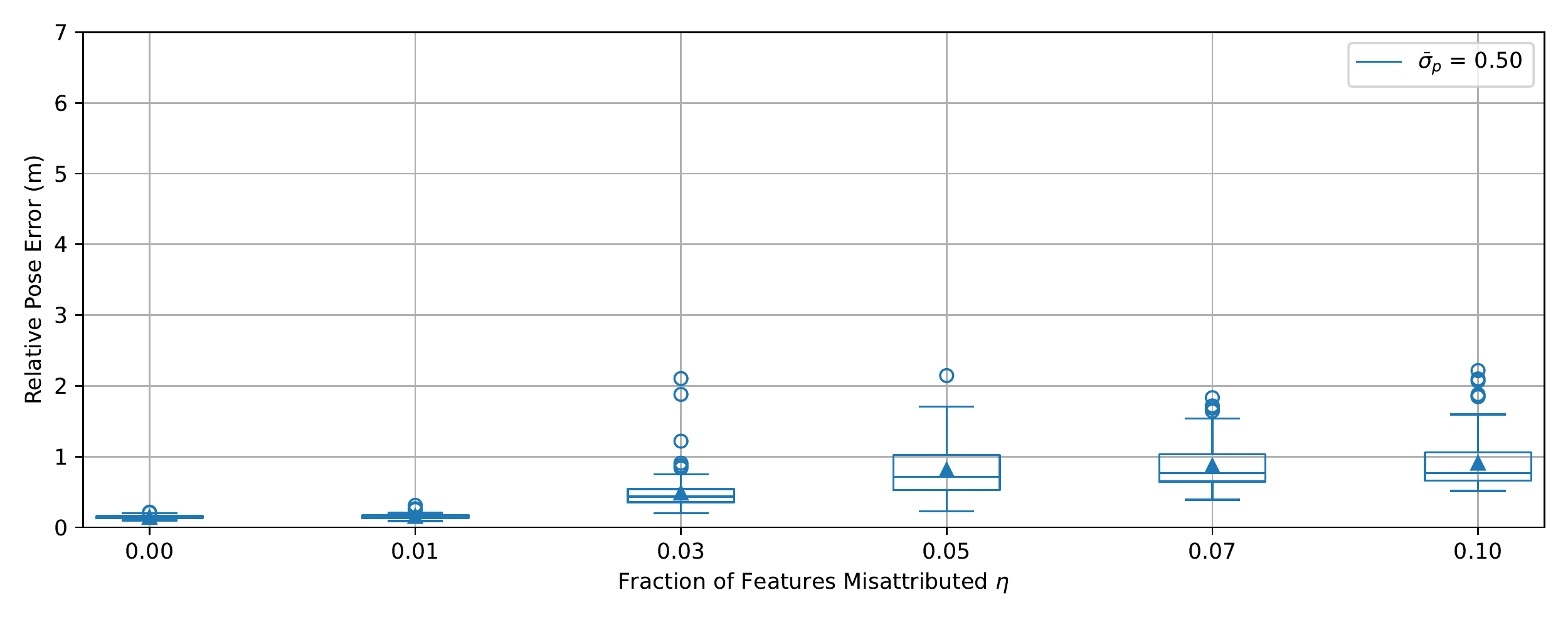}}
    \caption{\textbf{Attribution errors increase bias and variance in performance.}  Each box-and-whisker illustrates the distribution of ATE and RPE computed using 100 Monte-Carlo trials. Boxes extend from the first to the third quartile. Medians are lines in the boxes, means are triangles. Whiskers extend the box by 1.5x the inter-quartile range. }
    \label{fig:mismatch_performance}
\end{figure}

\begin{figure}
    \centering
    \includegraphics[width=6in]{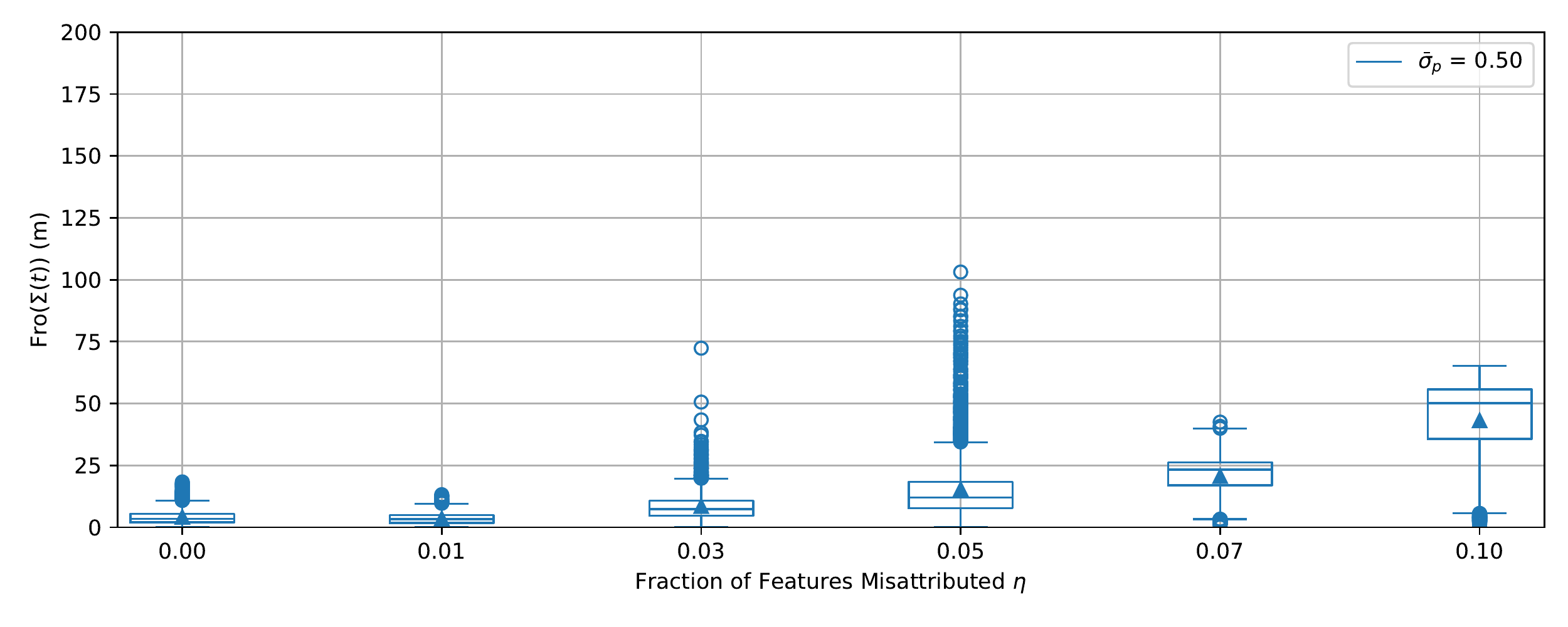}
    \caption{\textbf{Attribution errors produce more uncertainty.} Each box-and-whisker illustrates the distribution of mean sample covariance computed using 100 Monte-Carlo trials. Boxes extend from the first to the third quartile. Medians are lines in the boxes, means are triangles. Whiskers extend the box by 1.5x the inter-quartile range. }
    \label{fig:mismatch_covariance_norm}
\end{figure}

\begin{figure}
    \centering
    \includegraphics[width=6in]{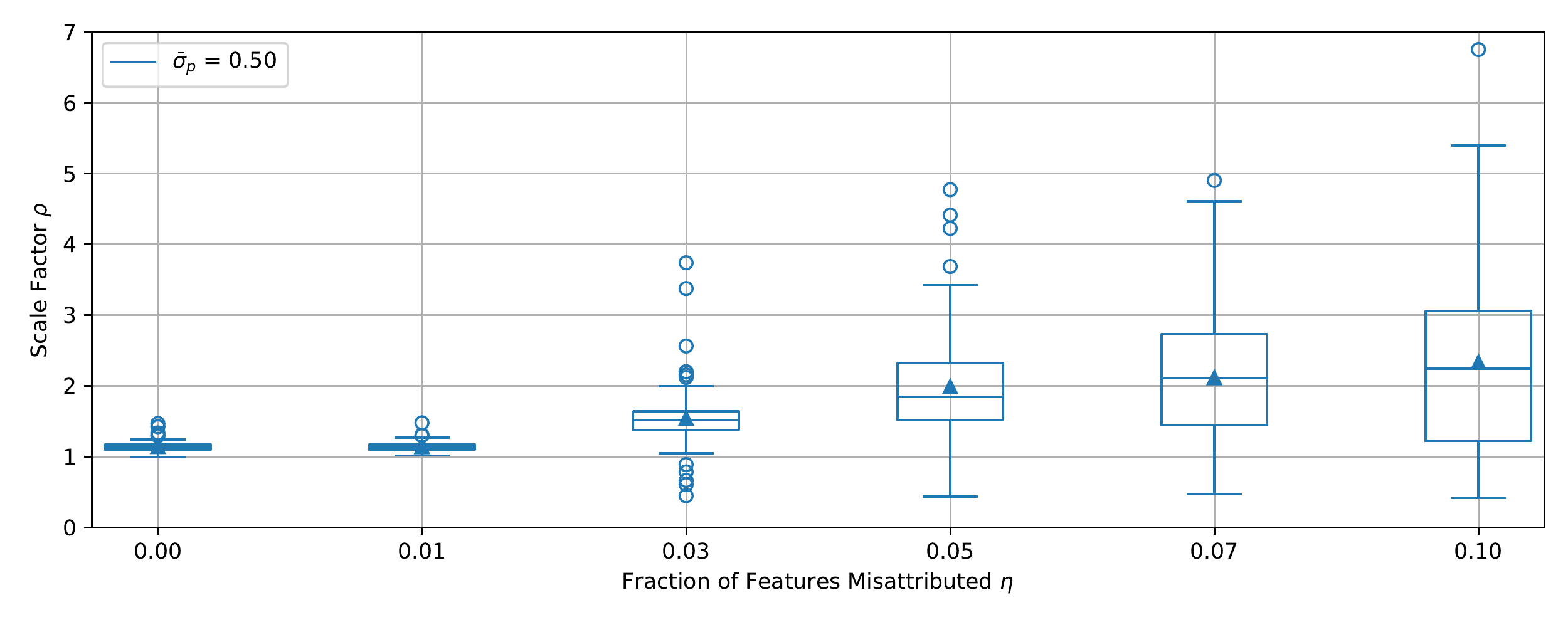}
    \caption{\textbf{Attribution errors increase bias and variance in scale.} Each box-and-whisker illustrates the distribution of scale factor computed using 100 Monte-Carlo trials. Boxes extend from the first to the third quartile. Medians are lines in the boxes, means are triangles. Whiskers extend the box by 1.5x the inter-quartile range.}
    \label{fig:mismatch_scale_distributions}
\end{figure}

The outputs of our Monte-Carlo experiments are the box-and-whisker plots of distributions of ATE, RPE, mean sample covariance, and scale factor, shown in Figures \ref{fig:gaussian_noise_performance} - \ref{fig:mismatch_scale_distributions}. As expected, increasing Gaussian noise, drift, and attribution errors increase ATE, RPE, mean sample covariance, and scale errors. The degradation with respect to Gaussian noise and drift is graceful. However, the degradation with respect to attribution errors is exponential. Cutting the lifetime of feature tracks causes the largest performance errors, mean sample covariance, scale bias, and scale variance. Next, Gaussian noise causes larger performance errors, mean sample covariance, and scale uncertainty than drift. Drift, however, causes larger scale bias than Gaussian noise.

More particular details are noted in the paragraphs below. All plots for the same quantity (e.g. distribution of $\rho$) use the same vertical-axis scale for the Gaussian noise and drift experiments. The plots containing results of the attribution experiments require a different vertical-axis scale because of the exponential degradation.

\paragraph{Gaussian Noise.} Results are shown in Figures \ref{fig:gaussian_noise_performance}, \ref{fig:gaussian_noise_covariance_norm}, \ref{fig:gaussian_noise_scale_distributions}. The general trend is that performance errors and state uncertainty, and variations of performance error and state uncertainty, increase with $\sigma_p$. The variation in $\rho$ generally increases with $\sigma_p$, but mean and median of $\rho$ hover around $2.1$.

Next, we note that for performance error and scale errors for $\sigma_p = 0.25$ are larger than performance error and scale errors for $\sigma_p = 0.50$. We hypothesize that the combination of frequent changes in motion that make the dynamics less linear and poor feature initialization add measurement errors that are not adequately captured when we set $\bar \sigma_p = \sigma_p$. When we instead set $\bar \sigma_p$ a little bit higher, to $\bar \sigma_p = \sigma_p + 0.25$, performance error and scale errors when $\sigma_p = 0.25$ become lower than when $\sigma_p = 0.50$. Results when $\bar \sigma_p = \sigma_p + 0.25$ are shown in Figures \ref{fig:gaussian_noise_performance2}, \ref{fig:gaussian_noise_covariance_norm2}, and \ref{fig:gaussian_noise_scale_distributions2}.

\paragraph{Drift.} Distributions of performance error, state covariance, and scale error in the drift experiment are shown in Figures \ref{fig:drift_performance}, \ref{fig:drift_covariance_norm}, \ref{fig:drift_scale_distributions}. Performance error, state covariance, and scale errors increase monotonically with $\sigma_b$ for both $\bar \sigma_p = 0.50$ and $\bar \sigma_p = 1.00$; their variations also widen as $\sigma_b$ increases. Values of performance error, state covariance, and $\rho$ are closer to their ideal values when $\bar \sigma_p = 0.50$ than when $\bar \sigma_p = 1.00$ for smaller quantities of drift. Once $\sigma_b$ is high enough, values of performance error, state covariance, and $\rho$ are closer to their ideal values with $\bar \sigma_p = 1.00$.

\begin{figure}
    \centering
    \includegraphics[width=0.48\textwidth]{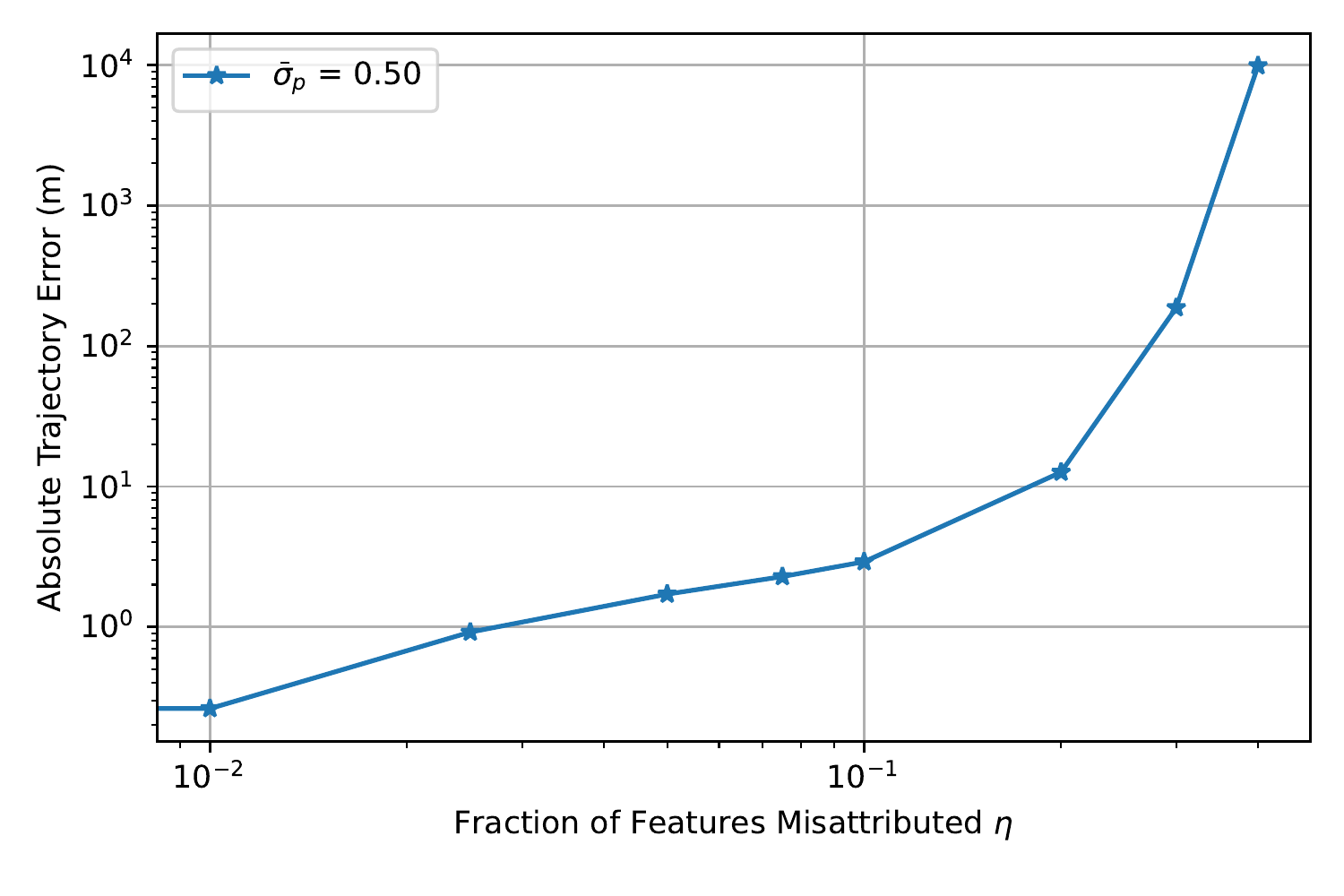}
    \includegraphics[width=0.48\textwidth]{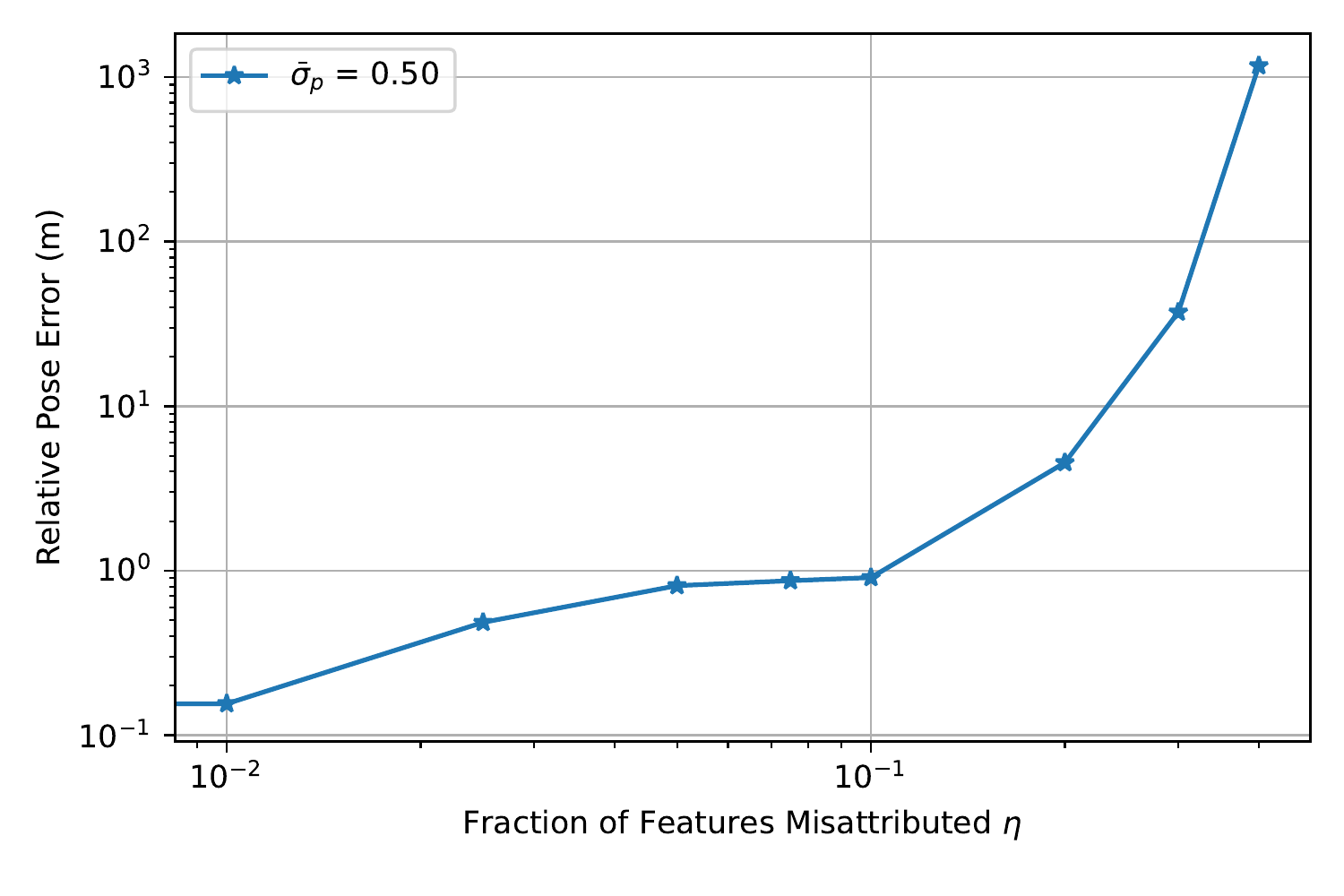}
    \includegraphics[width=0.48\textwidth]{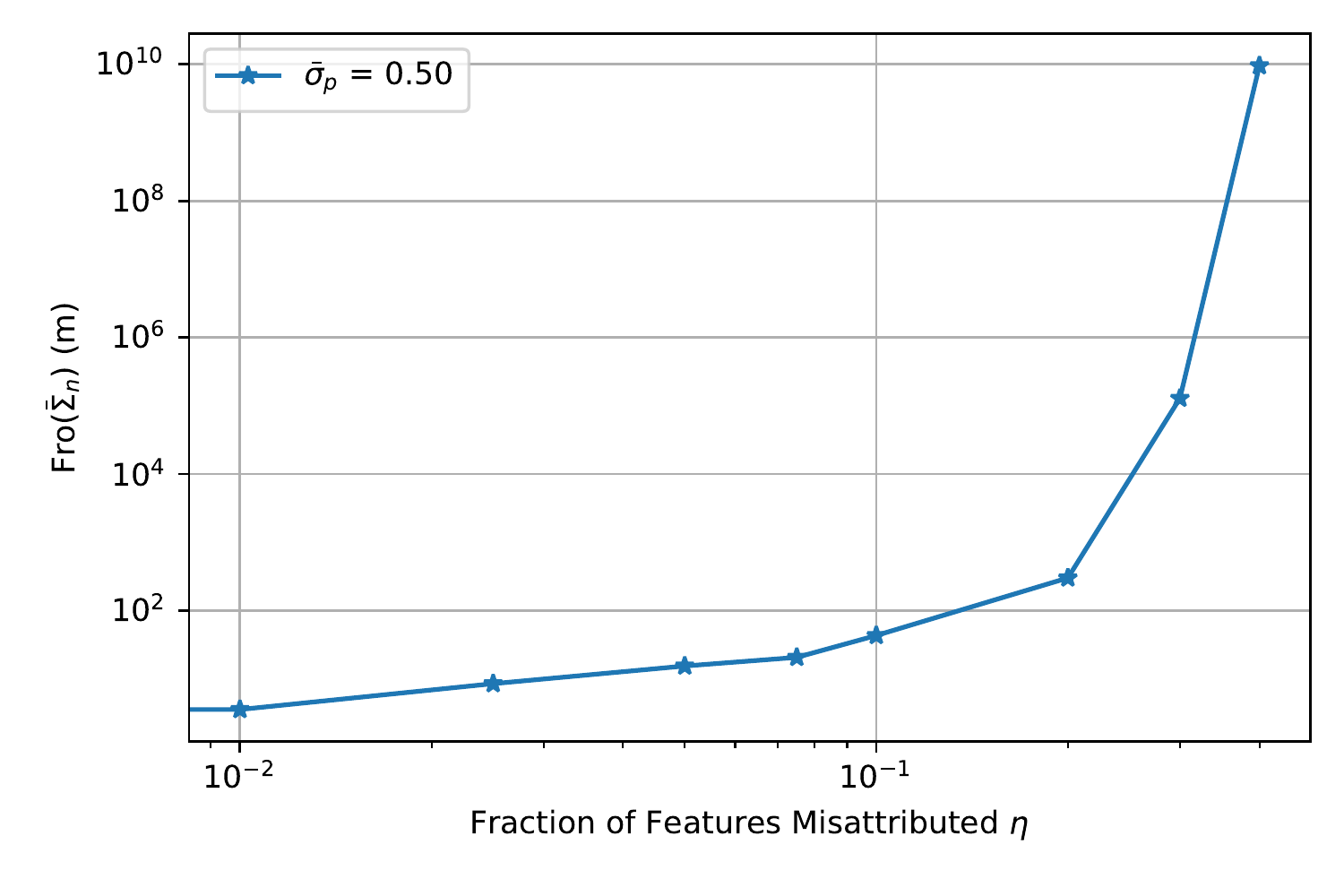}
    \includegraphics[width=0.48\textwidth]{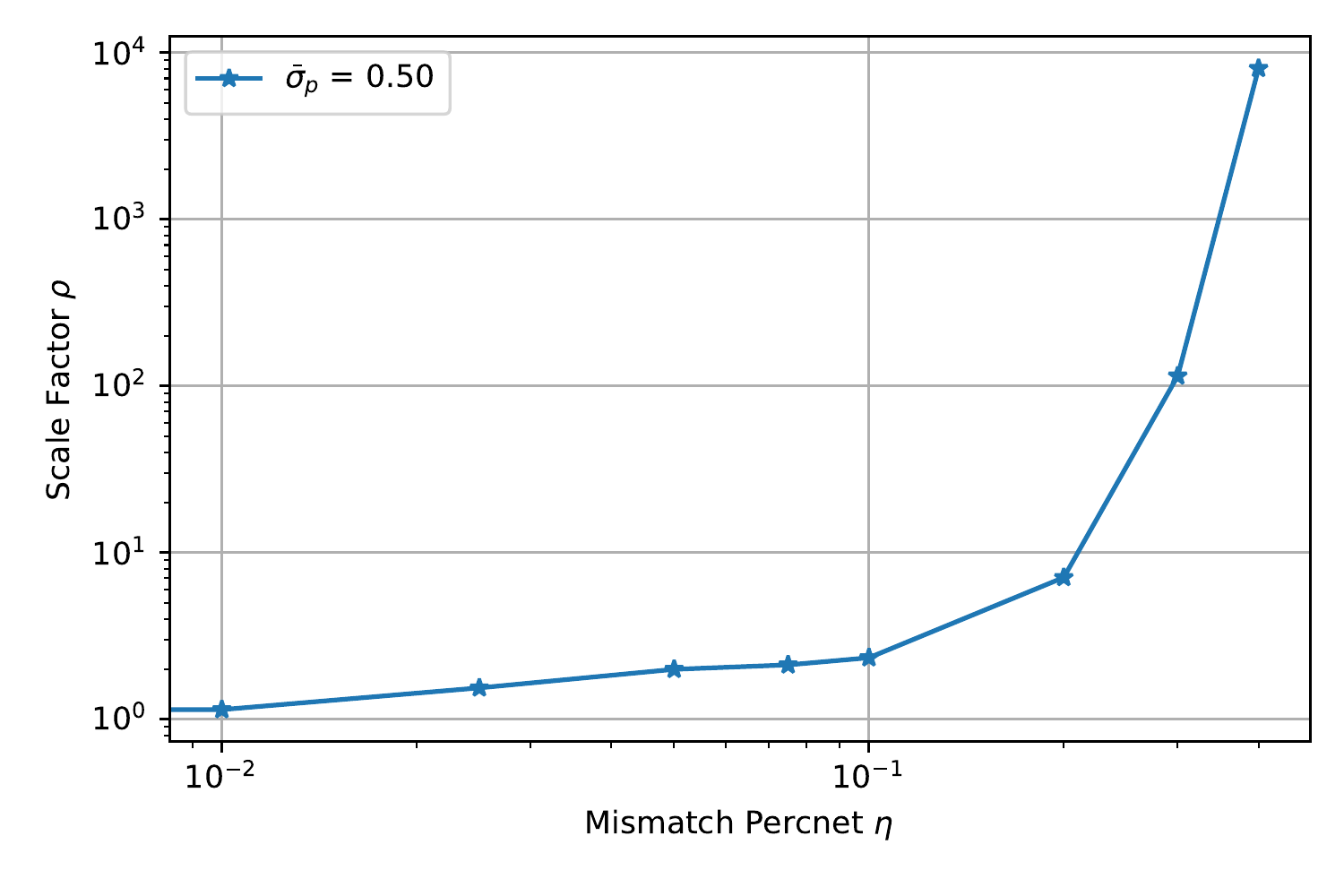}
    \caption{\textbf{The increases in performance errors, state uncertainty, and scale uncertainty due to attribution errors are exponential.} The four figures plot mean values of ATE, RPE, state uncertainty, and scale factor as a function of $\eta$, for $\eta \in [0.01, 0.025, 0.05, 0.075, 0.1, 0.2, 0.3, 0.4]$. Both the horizontal and vertical axes are in log scale. All curves are exponential functions of $\eta$.}
    \label{fig:eta_mean_line}
\end{figure}

\paragraph{Attribution Errors.}
Distributions of performance error, state covariance, and scale error are shown in Figures \ref{fig:mismatch_performance}, \ref{fig:mismatch_covariance_norm}, \ref{fig:mismatch_scale_distributions}. Results are simple: performance error, state covariance, and scale factor increase exponentially as $\eta$ is increased. Variation in all three increase uniformly with $\eta$. Figures \ref{fig:mismatch_performance}, \ref{fig:mismatch_covariance_norm}, and \ref{fig:mismatch_scale_distributions} only display values of $\eta$ up to $\eta = 0.1$, so that distributions of performance error, state covariance, and scale error at lower values are not dwarfed. Mean values of performance error, state covariance, and scale factor are shown for all tested values of $\eta$ in Figure \ref{fig:eta_mean_line}.

We note that $\eta = 0.1$ is a rather low percentage of outliers and that SLAM systems commonly encounter higher outlier ratios, as illustrated by the previous chapter. This indicates that outliers in real-world data are not random, like our simulated attribution errors, and that algorithms used to select features for state estimation are functioning as intended.

\section{Summary and Discussion}

There are at least two limitations to our results. The first is that we used theoretically correct values of $\bar \sigma_p$ in the Gaussian noise experiment rather than treating $\bar \sigma_p$ as a tuning parameter. An ideal experiment with infinite resources would find the very best\footnote{The notion of ``best'' also needs to be defined.} value of $\bar \sigma_p$ for every possible value of every disturbance and then compare the distribution of performance errors, mean sample covariance, and scales. The second limitation is that we studied a randomly generated trajectory to eliminate the possibility that the number of trajectories that could produce the same measurements could be unbounded. Many realistic trajectories, such as driving a car on a flat road or flying a drone at a constant velocity, are not sufficiently exciting. 

Nevertheless, we have systematically quantified and characterized the performance and uncertainty of an well-known monocular-VIO state estimation algorithm based on the Extended Kalman Filter. Our results largely confirm what is anecdotally known by practitioners: monocular VIO is ``finicky''. Whenever possible, a practitioner will always choose RGB-D SLAM, Lidar-Inertial Odometry, or Stereo Visual-Inertial Odometry if the platform allows it. The fliers in Figures \ref{fig:gaussian_noise_performance}, \ref{fig:gaussian_noise_scale_distributions}, \ref{fig:drift_performance}, and \ref{fig:drift_scale_distributions} indicate that the right combination of noise and/or drift in the IMU and visual measurements has more than a 1 in 100 chance of creating a state estimate with a performance error more than twice the mean error and far above the lowest possible error. This highlights the need for more Monte-Carlo studies of SLAM, rather than reliance on benchmark datasets with individual trajectories.

\bibliographystyle{plain}
\bibliography{xivo_features_calib/references.bib}

\begin{thebibliography}{10}

\bibitem{grewal_identifiability_1976}
M.~Grewal and K.~Glover.
\newblock Identifiability of linear and nonlinear dynamical systems.
\newblock {\em IEEE Transactions on Automatic Control}, 21(6):833--837,
  December 1976.
\newblock Conference Name: IEEE Transactions on Automatic Control.

\bibitem{hernandez_observability_2015}
J.~Hernandez, K.~Tsotsos, and S.~Soatto.
\newblock Observability, identifiability and sensitivity of vision-aided
  inertial navigation.
\newblock In {\em 2015 {IEEE} {International} {Conference} on {Robotics} and
  {Automation} ({ICRA})}, pages 2319--2325, May 2015.

\bibitem{jones_visual-inertial_2011}
E.~Jones and S.~Soatto.
\newblock Visual-inertial navigation, mapping and localization: {A} scalable
  real-time causal approach.
\newblock {\em The International Journal of Robotics Research}, 30(4):407--430,
  April 2011.

\bibitem{lavretsky_robust_2012}
Eugene Lavretsky and Kevin Wise.
\newblock {\em Robust and {Adaptive} {Control}: {With} {Aerospace}
  {Applications}}.
\newblock Springer Science \& Business Media, November 2012.
\newblock Google-Books-ID: a2128lhlWfQC.

\bibitem{Lee_2019_ICCV}
Seong~Hun Lee and Javier Civera.
\newblock Closed-form optimal two-view triangulation based on angular errors.
\newblock In {\em Proceedings of the IEEE/CVF International Conference on
  Computer Vision (ICCV)}, October 2019.

\bibitem{murray_mathematical_1994}
Richard~M. Murray, Zexiang Li, and S.~Shankar Sastry.
\newblock {\em A {Mathematical} {Introduction} to {Robotic} {Manipulation}}.
\newblock Routledge, Boca Raton, 1 edition edition, March 1994.

\bibitem{padoan_geometric_2017}
Alberto Padoan, Giordano Scarciotti, and Alessandro Astolfi.
\newblock A {Geometric} {Characterization} of the {Persistence} of {Excitation}
  {Condition} for the {Solutions} of {Autonomous} {Systems}.
\newblock {\em IEEE Transactions on Automatic Control}, 62(11):5666--5677,
  November 2017.
\newblock Conference Name: IEEE Transactions on Automatic Control.

\bibitem{soderstrom_identifiability_1976}
T.~Soderstrom, L.~Ljung, and I.~Gustavsson.
\newblock Identifiability conditions for linear multivariable systems operating
  under feedback.
\newblock {\em IEEE Transactions on Automatic Control}, 21(6):837--840,
  December 1976.
\newblock Conference Name: IEEE Transactions on Automatic Control.

\bibitem{sturm_benchmark_2012}
Jürgen Sturm, Nikolas Engelhard, Felix Endres, Wolfram Burgard, and Daniel
  Cremers.
\newblock A benchmark for the evaluation of {RGB}-{D} {SLAM} systems.
\newblock In {\em 2012 {IEEE}/{RSJ} {International} {Conference} on
  {Intelligent} {Robots} and {Systems}}, pages 573--580, October 2012.

\bibitem{tomei_enhanced_2022}
Patrizio Tomei and Riccardo Marino.
\newblock An {Enhanced} {Feedback} {Adaptive} {Observer} for {Nonlinear}
  {Systems} with {Lack} of {Persistency} of {Excitation}.
\newblock {\em IEEE Transactions on Automatic Control}, pages 1--6, 2022.
\newblock Conference Name: IEEE Transactions on Automatic Control.

\bibitem{tsuei_xivodocpdf_2023}
Stephanie Tsuei and Xiaohan Fei.
\newblock xivo/doc.pdf at devel · ucla-vision/xivo.
\newblock \url{https://github.com/ucla-vision/xivo/blob/devel/doc/doc.pdf},
  February 2023.

\bibitem{tsuei_feature_2023}
Stephanie Tsuei, Wenjie Mo, and Stefano Soatto.
\newblock Feature {Tracks} are not {Zero}-{Mean} {Gaussian}, March 2023.
\newblock arXiv:2303.14315 [cs].

\bibitem{tsuei_2021}
Stephanie Tsuei, Stefano Soatto, Paulo Tabuada, and Mark~B. Milam.
\newblock Learned uncertainty calibration for visual inertial localization.
\newblock In {\em 2021 IEEE International Conference on Robotics and Automation
  (ICRA)}, pages 5311--5317, 2021.

\bibitem{verrelli_nonanticipating_2020}
Cristiano~Maria Verrelli and Patrizio Tomei.
\newblock Nonanticipating {Lyapunov} {Functions} for {Persistently} {Excited}
  {Nonlinear} {Systems}.
\newblock {\em IEEE Transactions on Automatic Control}, 65(6):2634--2639, June
  2020.
\newblock Conference Name: IEEE Transactions on Automatic Control.

\bibitem{willems_note_2005}
Jan~C. Willems, Paolo Rapisarda, Ivan Markovsky, and Bart~L.M. De~Moor.
\newblock A note on persistency of excitation.
\newblock {\em Systems \& Control Letters}, 54(4):325--329, April 2005.

\bibitem{yang_degenerate_2019}
Yulin Yang, Patrick Geneva, Kevin Eckenhoff, and Guoquan Huang.
\newblock Degenerate {Motion} {Analysis} for {Aided} {INS} {With} {Online}
  {Spatial} and {Temporal} {Sensor} {Calibration}.
\newblock {\em IEEE Robotics and Automation Letters}, 4(2):2070--2077, April
  2019.
\newblock Conference Name: IEEE Robotics and Automation Letters.

\bibitem{astrom_numerical_1965}
Karl-Johan Åström and Bohlin Torsten.
\newblock Numerical {Identification} of {Linear} {Dynamic} {Systems} from
  {Normal} {Operating} {Records}.
\newblock {\em IFAC Proceedings Volumes}, 2(2):96--111, September 1965.

\end{thebibliography}

\end{document}